\documentclass{article}

\PassOptionsToPackage{numbers, compress}{natbib}
\usepackage[preprint]{neurips_2019}

\usepackage[utf8]{inputenc} % allow utf-8 input
\usepackage[T1]{fontenc}    % use 8-bit T1 fonts
\usepackage{booktabs}       % professional-quality tables
\usepackage{graphicx}
\usepackage{caption}
\usepackage{subcaption}
\usepackage{amsmath,amssymb,amsfonts}
\usepackage{nicefrac}
\usepackage{microtype}    
\usepackage{hyperref}

\usepackage{xspace}

\title{Detecting semantic anomalies}

\author{
  Faruk Ahmed\thanks{Mila, Université de Montréal. E-mail: faruk.ahmed@umontreal.ca} \\
  \And
  Aaron Courville\thanks{Mila, Université de Montréal, CIFAR Fellow. E-mail: aaron.courville@umontreal.ca} \\
}

\begin{document}

\allowdisplaybreaks

\newcommand{\mnist}{{\small \textsc{MNIST}}\xspace}
\newcommand{\cifar}{{\small \textsc{CIFAR-10}}\xspace}
\newcommand{\svhn}{{\small \textsc{SVHN}}\xspace}
\newcommand{\stl}{{\small \textsc{STL-10}}\xspace}
\newcommand{\timagenet}{{\small \textsc{tiny-Imagenet}}\xspace}
\newcommand{\lsun}{{\small \textsc{LSUN}}\xspace}
\newcommand{\imagenet}{{\small \textsc{Imagenet}}\xspace}

\maketitle
\begin{abstract}
We critically appraise the recent interest in out-of-distribution (OOD) detection and question the practical relevance of existing benchmarks. While the currently prevalent trend is to consider different datasets as OOD, we argue that out-distributions of practical interest are ones where the distinction is semantic in nature for a specified context, and that evaluative tasks should reflect this more closely. Assuming a context of object recognition, we recommend a set of benchmarks, motivated by practical applications. We make progress on these benchmarks by exploring a multi-task learning based approach, showing that auxiliary objectives for improved semantic awareness result in improved semantic anomaly detection, with accompanying generalization benefits. 
\end{abstract}

\section{Introduction}
In recent years, concerns have been raised about modern neural network based classification systems providing incorrect predictions with high confidence~\cite{guocalibration}. A possibly-related finding is that classification-trained CNNs find it much easier to ``overfit" to low-level properties such as texture~\cite{shapeiclr2019}, canonical pose~\cite{rotatedtruck}, or contextual cues~\cite{terraincognita} rather than learning globally coherent characteristics of objects.
A subsequent worry is that such classifiers, trained on data sampled from a particular distribution, are likely to be misleading when encountering novel situations in deployment. For example, silent failure might occur due to equally confident categorization of unknown objects into known categories (due to shared texture, for example).
%, due to having overfit to local biases.
This last concern is one of the primary motivating reasons for wanting to be able to detect when test data comes from a different distribution than that of the training data. This problem has been recently dubbed  \emph{out-of-distribution (OOD) detection}~\cite{amodei,danbaseline}, but is also referred to as anomaly/novelty/outlier detection in the contemporary machine learning context. Evaluation is typically carried out with benchmarks of the style proposed in~\cite{danbaseline}, where different datasets are treated as OOD after training on a particular in-distribution dataset. This area of research has been steadily developing, with some additions of new OOD datasets to the evaluation setup~\cite{odin}, and improved results. \\ \\
\textbf{Current benchmarks are ill-motivated} Despite such tasks rapidly becoming the standard benchmark for OOD detection in the community, we suggest that, taken as a whole, they are not very well-motivated. For example, the object recognition dataset \cifar (consisting of images of objects placed in the foreground), is typically trained and tested against noise, or different datasets such as downsampled \lsun (a dataset of scenes), or \svhn (a dataset of house numbers), or \timagenet (a different dataset of objects). For the simpler cases of noise, or datasets with scenes or numbers, low-level image statistics are sufficient to tell them apart. While choices like \timagenet might seem more reasonable, it has been noted that particular datasets have particular biases related to specific data collection and curation quirks~\cite{datasetbias,Tommasi2017}, which renders the problem of treating different datasets for OOD detection questionable. It is possible we are only getting better at distinguishing such idiosyncrasies. As an empirical illustration, we show in Appendix D that very trivial baselines can perform reasonably well at existing benchmarks. \\ \\
\textbf{\emph{Semantic} distributional shift is relevant} We call into question the practical relevance of these evaluative tasks which are currently treated as standard by the community. While they might have some value as very preliminary reliability certification or as a testbed for diagnosing peculiar pathologies (for example, undesired behaviours of unsupervised density models, as in~\cite{glowfail}), their significance as benchmarks for practical OOD detection is less clear. The implicit goal for the current style of benchmarks is that of detecting one or more of a wide variety of distributional shifts, which mostly consist of irrelevant factors when high-dimensional data has low-dimensional semantics. We argue that this is misguided; in a realistic setting, distributional shift across non-semantic factors (for example, camera and image-compression artefacts) is something we want to be robust to, while shift in semantic factors (for example, object identity) should be flagged down as anomalous or novel. Therefore, OOD detection is well-motivated only when the distributional shift is semantic in nature. \\ \\
\textbf{Context determines semantic factors} In practical settings, OOD detection becomes meaningful only after acknowledging context, which identifies relevant semantic factors of interest. These are the factors of variation whose unnatural deviation are of concern to us in our assumed context. For example, in the context of scene classification, a kitchen with a bed in the middle is an anomalous observation. However, in the context of object recognition, the primary semantic factor is not the composition of scene-components anymore, but the identity of the foreground object. Now the unusual context should not prevent correct object recognition. If we claim that our object recognition models should be less certain of identifying an object in a novel context, it amounts to saying that we would prefer our models to be biased. In fact, we would like our models to systematically generalize~\cite{fodor} in order to be trustworthy and useful. We would like them to form predictions from a globally coherent assimilation of the relevant semantic factors for the task, while being robust to their composition with non-semantic factors. \\ \\
\textbf{Without context, OOD detection is too broad to be meaningful} The problem of OOD detection then, as currently treated by the community, suffers from imprecision due to context-free presumption and evaluation. Even though most works assume an underlying classification task, the benchmark OOD datasets include significant variation over non-semantic factors. OOD detection with density models are typically presented as being unaware of a downstream module, but we argue that such a context must be specified in order to determine what shifts are of concern to us, since we typically do not care about all possible variations. Being agnostic of context when discussing OOD detection leads to a corresponding lack of clarity about the implications of underlying methodologies in proposed approaches. The current benchmarks and methods therefore carry a risk of potential misalignment between evaluative performance and field performance in practical OOD detection problems. Henceforth, we shall refer to such realistic OOD detection problems, where the concerned distributional shift is a semantic variation for a specified context, by the term \emph{anomaly detection}. \\ \\
\textbf{Contributions and overview} Our contributions in this paper are summarized as follows.

1. \emph{Semantic shifts are interesting, and benchmarks should reflect this more closely:} We provided a grounded discussion about the relevance of semanticity in the context of a task for realistic OOD (anomaly) detection. Under the view of regarding distributional shifts as being either semantic or non-semantic for a specified context, we concluded that semantic shifts are of practical interest. If we want to deploy reliable models in the real world, we typically wish to achieve robustness against non-semantic shift.

2. \emph{More practical benchmarks for anomaly detection:} Although our discussion applies generally, in this paper we assume the common context of object recognition. In this context, unseen object categories may be considered anomalous at the ``highest level'' of semanticity. Anomalies corresponding to intermediate levels of semantic decomposition can also be relevant; for example, a liger should result in 50-50 uncertainties if the training data contains only lions and tigers. However, such anomalies are significantly harder to curate, requiring careful interventions at collection-time. Since detection of novel categories is a compelling anomaly detection task in itself, we recommend benchmarks that reflect such applications in section 2.

3. \emph{Auxiliary objectives for improved semantic representation improves anomaly detection:} Following our discussion about the relevance of semanticity, in sections 4 and 5 we investigate the effectiveness of multi-task learning with auxiliary self-supervised objectives. These have been shown to result in semantic representations, measured through linear separability by object categories. Our experimental results are indicative that such augmented objectives lead to improved anomaly detection, with accompanying improvements in generalization.

\begin{table*}[t]
\caption{\small Sizes of proposed benchmark subsets from ILSVRC2012. Sample images are in the Appendix. The training set consists of roughly 1300 images per member, and 50 images per member in the test set (which come from the validation set images in the ILSVRC2012 dataset).}
\centering
\small
\begin{tabular}{@{}lccc@{}}
\toprule
Subset & Number of members & Total training images & Total test images \\
\midrule
Dog (\emph{hound dog}) & 12 & 14864 & 600	\\
Car & 10 & 13000 & 500 \\
Snake (\emph{colubrid snake}) &	9 & 11700 & 450 \\
Spider & 6 & 7800 & 300	\\
Fungus & 6 & 7800 & 300	\\
\bottomrule
\end{tabular}
\label{tab:benchmark}
\end{table*}

\section{Motivation and proposed tasks}
In order to develop meaningful benchmarks, we begin by considering some practical applications where being able to detect anomalies, in the context of classification tasks, would find use.

\emph{Nature studies and monitoring:} Biodiversity scientists want to keep track of variety and statistics of species across the world. Online tools such as \emph{iNaturalist} \cite{inaturalistapp} enable photo-based classification and subsequent cataloguing in data repositories from pictures uploaded by naturalists. In such automated detection tools, a potentially novel species should result in a request for expert help rather than misclassification into a known species, and detection of undiscovered species is in fact a task of interest. A similar practical application is camera-trap monitoring of members in an ecosystem, notifying caretakers upon detection of invasive species~\cite{pests,willi}. 
Taxonomy of collected specimens is often backlogged due to the human labour involved. Automating digitization and identification can help catch up, and often new species are brought to light through the process~\cite{historical}, which obviously depends on effective detection of novel specimens.

\emph{Medical diagnosis and clinical microbiology:} Online medical diagnosis tools such as \emph{Chester}~\cite{chester} can be impactful at improving healthcare levels worldwide. Such tools should be especially adept at being able to know when faced with a novel pathology rather than categorizing into a known subtype. Similar desiderata applies to being able to quickly detect new strains of pathogens when using machine learning systems to automate clinical identification in the microbiology lab~\cite{plosone}.

\emph{AI safety:}~\cite{amodei} discuss the problem of distributional shift in the context of autonomous agents operating in our midst, with examples of actions that do not translate well across domains. A similar example in that vein, grounded in a computer vision classification task, is the contrived scenario of encountering a novel vehicle (that follows different dynamics of motion), which might lead to a dangerous decision by a self-driving car which fails to recognize unfamiliarity.

Having compiled the examples above, we can now try to come up with an evaluative setting more aligned with realistic applications. The basic assumptions we make about possible evaluative tasks are: (i) that anomalies of practical interest are semantic in nature; (ii) that they are relatively rare events whose detection is of more primary relevance than minimizing false positives; and (iii) that we do not have access to examples of anomalies. These assumptions guide our choice of benchmarks  and evaluation. \\ \\
\textbf{Recommended benchmarks} A very small number of recent works~\cite{ganomaly,zenati} have considered a case that is more aligned with the goals stated above. Namely, for a choice of dataset, for example \mnist, train as many versions of classifiers as there are classes, holding out one class every time. At evaluation time, score the ability of being able to detect the held out class as anomalous. This is a setup more clearly related to the task of being able to detect semantic anomalies, holding dataset-bias factors invariant to a significantly greater extent. 
%Holding out each class at a time captures greater variance of detection performance than evaluation for one split of categories into normal/anomalous sets. %% I already say this later
In this paper, we shall explore this setting with \cifar and \stl, and recommend this as the default benchmark for evaluating anomaly detection in the context of object recognition. Similar setups apply to different contexts. We discourage the recently-adopted practice of treating one category as in-distribution and many other categories as out-distributions (as in~\cite{gpnd,golan}, for example). While this setting is not aligned with the context of multi-object classification, it relies on a dataset constructed for such a purpose. Moreover, practical situations calling for one-class modelling typically consider anomalies of interest to be (often subtle) variations of the same object, and not a set of very distinct categories.

While the hold-out-class setting for \cifar and \stl is a good setup for testing anomaly detection of disparate objects, a lot of applications, including some of the ones we described earlier, require detection of more fine-grained anomalies. For such situations, we propose a suite of tasks comprised of subsets of {\small \textsc{ILSVRC2012}}~\cite{imagenet2012}, with fine-grained subcategories. For example, the \textsc{spider} subset consists of members \emph{tarantula}, \emph{Argiope aurantia}, \emph{barn spider}, \emph{black widow}, \emph{garden spider}, \emph{wolf spider}. We also propose \textsc{fungus}, \textsc{dog}, \textsc{snake}, and \textsc{car} subsets. These subsets have varied sizes, with some of them being fairly small (see Table~\ref{tab:benchmark}). Although this is a significantly harder task, we believe this setting aligns with the practical situations we described above, where sometimes large quantities of labelled data are not always available, and a particular fine-grained selection of categories is of interest. See Appendix A for more details about our construction. \\ \\
\textbf{Evaluation} Current works tend to mainly use both Area under the Receiver-Operator-Characteristics (AUROC) and Area under Precision-Recall curve (AUPRC) to evaluate performance on anomaly detection. In situations where positive examples are not only much rarer, but also of primary interest for detection, AUROC scores are a poor reflection of detection performance; \emph{precision} is more relevant than the false positive rate~\cite{rocfawcett,prroc,Avati2018}. We shall not inspect AUROC scores because in all of our settings, normal examples significantly outnumber anomalous examples, and AUROC scores are insensitive to skew, thus resulting in optimistic scores~\cite{prroc}. Precision and recall are calculated as
\begin{eqnarray}
    \text{precision} &=& \frac{\text{true positives}}{\text{true positives + false positives}}, \\
    \text{recall} &=& \frac{\text{true positives}}{\text{true positives + false negatives}},
\end{eqnarray}
and a precision-recall curve is then defined as a set of precision-recall points 
\begin{eqnarray}
    \text{PR curve} \triangleq \{\text{recall}(t), \text{precision}(t), -\infty < t  < \infty\},
\end{eqnarray}
where $t$ is a threshold parameter.

The area under the precision-recall curve is calculated by varying the threshold $t$ over a range spanning the data, and creating a finite set of points for the PR curve. One alternative is to interpolate these points, producing a continuous curve as an approximation to the true curve, and computing the area under the interpolation by, for example, the trapezoid rule. Interpolation in a precision-recall curve can sometimes be misleading, as studied in~\cite{averageprecision}, who recommend a number of more robust estimators. Here we use the standard approximation to average precision as the weighted mean of precisions at thresholds, weighted by the increase in recall from the previous threshold
\begin{eqnarray}
    \text{average precision} = \sum_k \text{precision}_k ( \text{recall}_k - \text{recall}_{k-1}).
\end{eqnarray}
\section{Related work}
\textbf{Evaluative tasks} As discussed earlier, the style of benchmarks widely adopted today follows the recommendation in~\cite{danbaseline}. Among follow-ups, the most significant successor has been~\cite{odin} which augmented the suite of tests with slightly more reasonable choices: for example, \timagenet is considered as out-of-distribution for in-distrbution datasets such as \cifar. However, on closer inspection, we find that \timagenet shares semantic categories with \cifar, such as species of \{dogs, cats, frogs, birds\}, so it is unclear how such choices of evaluative tasks correspond to realistic anomaly detection problems. Work in the area of \emph{open-set recognition} is closer to a realistic setup in terms of evaluation; in~\cite{openset2}, detection of novel categories is tested with a set of images corresponding to different classes that were discontinued in subsequent versions of Imagenet, but later work~\cite{openset4} relapsed into treating very different datasets as novel. We do not encourage using one particular split of a collection of unseen classes as anomalous. This is because such a one-time split might favour implicit biases in the predefined split, and the chances of this happening is reduced with multiple hold-out trials.  As mentioned earlier, a small number of works have already used the hold-out-class style of tasks for evaluation. Unfortunately, due to a lack of a motivating discussion, the community at large continues to adopt the tasks in~\cite{danbaseline} and~\cite{odin}. \\ \\
\textbf{Approaches to OOD detection} In~\cite{danbaseline}, the most natural baseline for a trained classifier is presented, where the detection score is simply given by the predictive confidence of the classifier (MSP). Follow-up work in~\cite{odin} proposed adding a small amount of adversarial perturbation, followed by temperature scaling of the softmax (ODIN). Methodologically, the approach suffers from having to pick a temperature and perturbation weight per anomaly-dataset. Complementary methods such as confidence calibration of~\cite{taylor}, have been shown to improve performance of MSP and ODIN.

Using auxiliary datasets as surrogate anomalies has been shown to improve performance on existing benchmarks in~\cite{oe}. This approach is limited, due to its reliance on other datasets, but a more practical variant in~\cite{kimin} uses a GAN to generate negative samples. However,~\cite{kimin} suffers from the methodological issue of hyperparameters being optimized per anomaly-dataset. We believe that such contentious practices arise from a lack of a clear discussion of the nature of the tasks we should be concerned with, and a lack of grounding in practical applications which would dictate proper methodology. The primary goal of our paper is to help fill this gap.

In \cite{multiplelabels}, the training set is augmented with semantically similar labels, but it is not always practical to assume access to a corpora providing such labels. In the next part of the paper, we explore a way to potentially induce more semantic representation, with the hope that this would lead to corresponding improvements in semantic anomaly detection and generalization.
\begin{table}[!t]
\caption{\small Multi-task augmentation with the self-supervised objective of predicting rotation improves generalization.}
\small
\centering
\begin{tabular}{@{}p{4.2cm}cc@{}} \toprule
   & \cifar & \stl \\
  \midrule
  Classification only & $95.87 \pm 0.05$  & $85.51 \pm 0.17$ \\
  Classification+rotation & $96.54 \pm 0.08$ & $88.98 \pm 0.30 $  \\
  \bottomrule
  \end{tabular}
\label{table:rotationhelps}%
\end{table}
\begin{figure*}[t]
    \centering
    \includegraphics[scale=0.25]{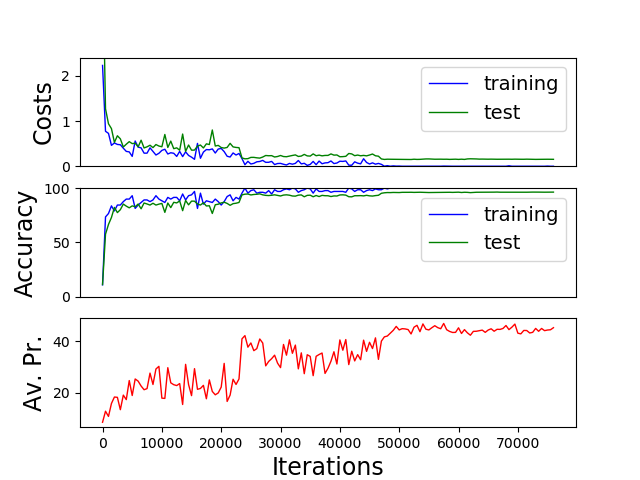}
    \includegraphics[scale=0.25]{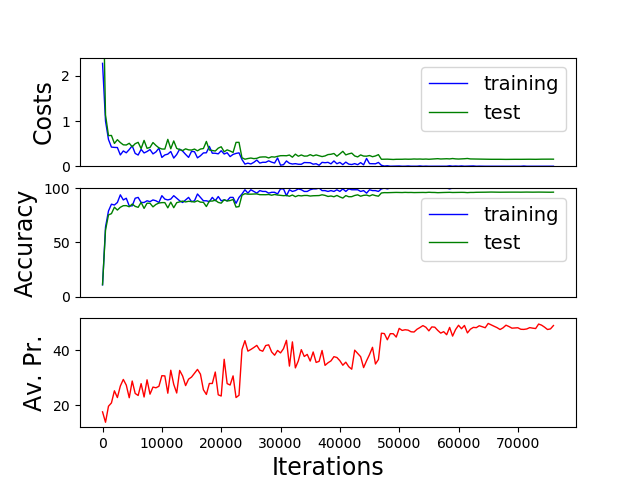}
    \includegraphics[scale=0.25]{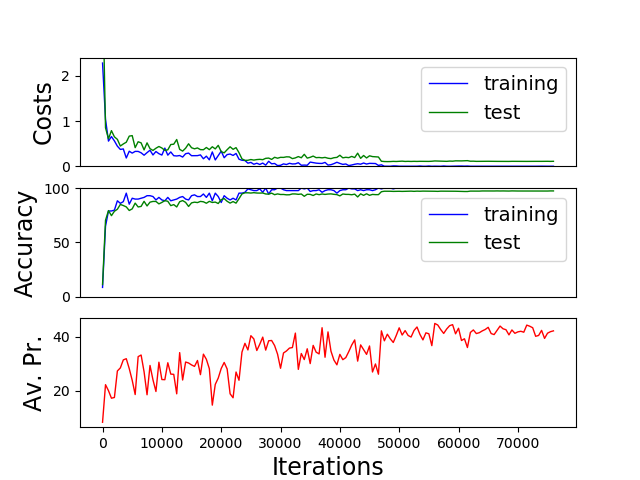} \\
    \includegraphics[scale=0.25]{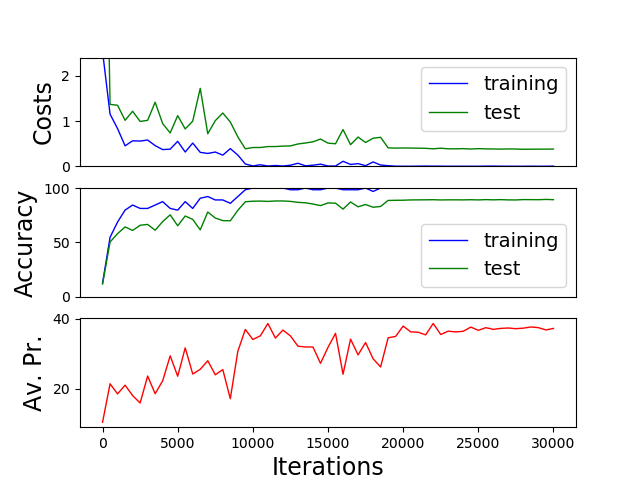}
    \includegraphics[scale=0.25]{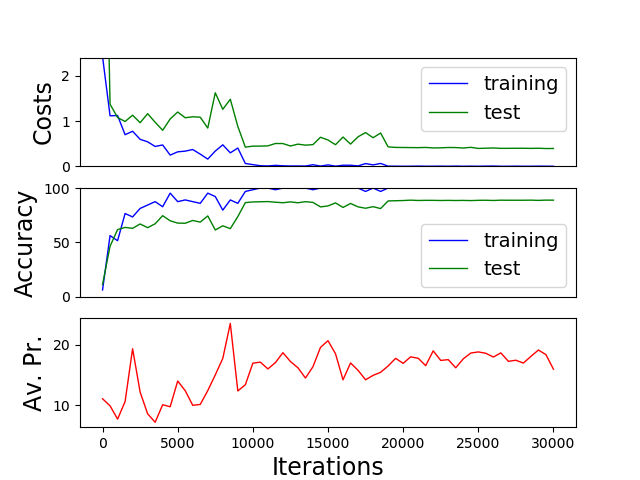}
    \includegraphics[scale=0.25]{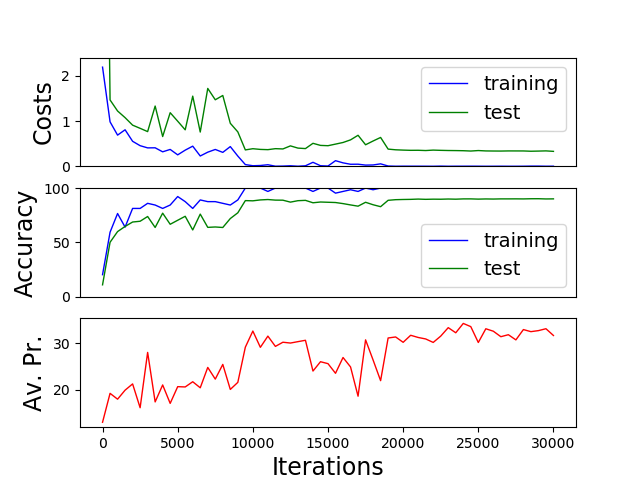}
    \caption{\small Plots of costs, accuracies, and average precision for hold-out-class experiments with 3 categories each from CIFAR-10 (top) and STL-10 (bottom), using the MSP method~\cite{danbaseline}. While classification performance is not correlated with performance at anomaly detection (compare test accuracy numbers with average precision scores), the ``pattern'' of improvement at anomaly detection appears roughly related to generalization (compare the coarse shape of test accuracy curves with that of average precision curves).}
    \label{fig:corr}
\end{figure*}
\section{Encouraging semantic representations with auxiliary self-supervised objectives}
We hypothesize that classifiers that learn representations which are more oriented toward capturing semantic properties would naturally lead to better performance at detecting semantic anomalies. ``Overfitting" to low-level features such as colour or texture without consideration of global coherence might result in potential confusions in situations where the training data is biased and not representative. For a lot of existing datasets, it is quite possible to achieve good generalization performance without learning semantic distinctions, a possibility that spurs the search for removing algorithmic bias~\cite{fair}, and which is often exposed in embarrassing ways. As a contrived example, if the training and testing data consists of only one kind of animal which is furry, the classifier only needs to learn about fur-texture, and can ignore other meaningful characteristics such as the shape. Such a system would fail to recognize another furry, but differently shaped creature as novel, while achieving good test performance. Motivated by this line of thinking, we ask the question of how we might encourage classifiers to learn more meaningful representations. \\ \\
\textbf{Multi-task learning with auxiliary objectives} Caruana~\cite{montreal} describes how sharing parameters for learning multiple tasks, which are related in the sense of requiring similar features, can be a powerful tool for inducing domain-specific inductive biases in a learner. Hand-design of inductive biases requires complicated engineering, while using the training signal from a related task can be a much easier way to achieve similar goals. Even when related tasks are not explicitly available, it is often possible to construct one. We explore such a framework for augmenting object recognition classifiers with auxiliary tasks. Expressed in notation, given the primary loss function, $\ell_{\text{primary}}$, which is the categorical cross-entropy loss in the case of classification, and the auxiliary loss $\ell_{\text{auxiliary}}$ corresponding to the auxiliary task, we aim to optimize the combined loss
\begin{equation}
    \ell_{\text{combined}}(\theta; \mathcal{D}) = \ell_{\text{primary}}(\theta; \mathcal{D}) + \lambda \ell_{\text{auxiliary}}(\theta; \mathcal{D}),
\end{equation}
where $\theta$ are the shared parameters across both tasks, $\mathcal{D}$ is the dataset, $\lambda$ is a hyper-parameter we learn by optimizing for classification accuracy on the validation set. In practice, we alternate between the two updates rather than taking one global step; this balances the \emph{training  rates} of the two tasks. \\ \\
\textbf{Auxiliary tasks} Recently, there has been strong interest in self-supervision applied to vision~\cite{doersch2015,pathak2016,noroozi2016,zhang2017,cpc,gidaris2018,cluster}, exploring tasks that induce representations which are linearly separable by object categories. These objectives naturally lend themselves as auxiliary tasks for encouraging inductive biases towards semantic representations. First, we experiment with the recently introduced task in~\cite{gidaris2018}, which asks the learner to predict the orientation of a rotated image. In Table~\ref{table:rotationhelps}, we show significantly improved generalization performance of classifiers on \cifar and \stl when augmented with the auxiliary task of predicting rotation. Details of experimental settings, and performance on anomaly detection, are in the next section. We also perform experiments on anomaly detection with contrastive predictive coding~\cite{cpc} as the auxiliary task and find that similar trends continue to hold.

The addition of such auxiliary objectives is complementary to the choice of scoring anomalies. Additionally, it enables further augmentation with more auxiliary tasks~\cite{mltselfsup}. 
\section{Evaluation}
We study the two existing representative baselines of maximum softmax probability (MSP)~\cite{danbaseline},  and ODIN~\cite{odin} on the proposed benchmarks. For ODIN, it is unclear how to choose the hyperparameters for temperature scaling and the weight for adversarial perturbation without assuming access to anomalous examples, an assumption we consider unrealistic in most practical settings. We fix $T = 1000, \epsilon = 5\text{e-}5$ for all experiments, following the most common setting.
\begin{table*}
\caption{\small We train ResNet classifiers on \cifar holding out each class per run, and score detection with average precision for the maximum softmax probability (MSP) baseline in~\cite{danbaseline}  and ODIN~\cite{odin}. We find that augmenting with rotation results in improved anomaly detection as well as generalization (contrast columns in the right half with the left).}
\centering
\small
\scalebox{0.95}{
\begin{tabular}{@{}p{1.3cm}cccccccc@{}}
\toprule
\textit{CIFAR-10} & \multicolumn{3}{c}{Classification-only} & \multicolumn{3}{c}{Rotation-augmented} \\
\cmidrule(lr){2-4}
\cmidrule(lr){5-7}
Anomaly & MSP & ODIN & Accuracy & MSP & ODIN & Accuracy \\
\midrule
airplane & 43.30 $\pm$ 1.13 & 48.23 $\pm$ 1.90 & 96.00 $\pm$ 0.16 & 46.87 $\pm$ 2.10 & 49.75 $\pm$ 2.30 & 96.91 $\pm$ 0.02 \\
automobile & 14.13 $\pm$ 1.33 & 13.47 $\pm$ 1.50 & 95.78 $\pm$ 0.12 & 17.39 $\pm$ 1.26 & 17.35 $\pm$ 1.12 & 96.66 $\pm$ 0.03 \\
bird & 46.55 $\pm$ 1.27 & 50.59 $\pm$ 0.95 & 95.90 $\pm$ 0.17 & 51.49 $\pm$ 1.07 & 54.62 $\pm$ 1.10 & 96.79 $\pm$ 0.06 \\
cat & 38.06 $\pm$ 1.31 & 38.97 $\pm$ 1.43 & 97.05 $\pm$ 0.12 & 53.12 $\pm$ 0.92 & 55.80 $\pm$ 0.76 & 97.46 $\pm$ 0.07 \\
deer & 49.11 $\pm$ 0.53 & 53.03 $\pm$ 0.50 & 95.87 $\pm$ 0.12 & 50.35 $\pm$ 2.57 & 52.82 $\pm$ 2.96 & 96.76 $\pm$ 0.09 \\
dog & 25.39 $\pm$ 1.17 & 24.41 $\pm$ 1.05 & 96.64 $\pm$ 0.13 & 32.11 $\pm$ 0.82 & 32.46 $\pm$ 1.39 & 97.36 $\pm$ 0.06 \\
frog & 40.91 $\pm$ 0.81 & 42.21 $\pm$ 0.48 & 95.65 $\pm$ 0.09 & 52.39 $\pm$ 4.58 & 54.44 $\pm$ 5.80 & 96.51 $\pm$ 0.12 \\
horse & 36.18 $\pm$ 0.77 & 36.78 $\pm$ 0.82 & 95.64 $\pm$ 0.08 & 39.93 $\pm$ 2.30 & 39.65 $\pm$ 4.31 & 96.27 $\pm$ 0.07 \\
ship & 28.35 $\pm$ 0.81 & 30.61 $\pm$ 1.46 & 95.70 $\pm$ 0.15 & 29.36 $\pm$ 3.16 & 28.82 $\pm$ 4.63 & 96.66 $\pm$ 0.17 \\
truck & 27.17 $\pm$ 0.73 & 28.01 $\pm$ 1.06 & 96.04 $\pm$ 0.24 & 29.22 $\pm$ 2.87 & 29.93 $\pm$ 3.86 & 96.91 $\pm$ 0.12 \\
\midrule
Average & 34.92 $\pm$ 0.41 & 36.63 $\pm$ 0.61 & 96.03 $\pm$ 0.00 & 40.22 $\pm$ 0.16 & 41.56 $\pm$ 0.15 & 96.83 $\pm$ 0.02 \\
\bottomrule
\end{tabular}
}
\label{tab:cifar}
\end{table*}
\begin{table*}[t]
\caption{\small Average precision scores for hold-out-class experiments with \stl. We observe that the same trends in improvements hold as with the previous experiments on \cifar.}
\small
\centering
\scalebox{0.95}{
\begin{tabular}{@{}p{1.3cm}cccccc@{}}
\toprule
\textit{STL-10}& \multicolumn{3}{c}{ Classification-only} & \multicolumn{3}{c}{Rotation-augmented}  \\
\cmidrule(lr){2-4}
\cmidrule(lr){5-7}
Anomaly & MSP & ODIN & Accuracy & MSP & ODIN & Accuracy \\
\midrule
airplane & 19.21 $\pm$ 1.05 & 23.46 $\pm$ 1.65 & 85.18 $\pm$ 0.20 & 22.21 $\pm$ 0.76 &	23.37 $\pm$ 1.71 & 89.24 $\pm$ 0.12 \\
bird & 29.05 $\pm$ 0.69 & 33.51 $\pm$ 0.36 & 85.91 $\pm$ 0.36 & 36.12 $\pm$ 2.08 & 40.08 $\pm$ 3.30 & 89.91 $\pm$ 0.29\\
car & 14.52 $\pm$ 0.37 & 16.14 $\pm$ 0.83 & 84.32 $\pm$ 0.55 & 15.95 $\pm$ 2.20 & 16.87 $\pm$ 2.94 & 89.52 $\pm$ 0.44 \\
cat & 25.21 $\pm$ 0.93 & 27.92 $\pm$ 0.84 & 86.95 $\pm$ 0.36 & 29.34 $\pm$ 1.30 & 31.35 $\pm$ 1.88 & 90.89 $\pm$ 0.26 \\
deer & 24.29 $\pm$ 0.53 & 25.94 $\pm$ 0.49 & 85.34 $\pm$ 0.35 & 27.60 $\pm$ 2.22 & 29.71 $\pm$ 2.55 & 89.20 $\pm$ 0.17 \\
dog & 23.42 $\pm$ 0.60 & 23.44 $\pm$ 1.18 & 87.78 $\pm$ 0.45 & 26.78 $\pm$ 0.71 & 26.14 $\pm$ 0.62 & 91.37 $\pm$ 0.33 \\
horse & 21.31 $\pm$ 1.01 & 22.19 $\pm$ 0.75 & 85.52 $\pm$ 0.21 & 23.79 $\pm$ 1.46 & 23.59 $\pm$ 1.63 & 89.60 $\pm$ 0.11 \\
monkey & 23.67 $\pm$ 0.83 & 21.98 $\pm$ 0.91 & 86.66 $\pm$ 0.31 & 28.43 $\pm$ 1.67 & 28.32 $\pm$ 1.20 & 90.07 $\pm$ 0.23 \\
ship & 14.61 $\pm$ 0.12 & 13.78 $\pm$ 0.63 & 84.65 $\pm$ 0.21 & 16.79 $\pm$ 1.20 & 15.37 $\pm$ 1.22 & 89.33 $\pm$ 0.15 \\
truck & 15.43 $\pm$ 0.17 & 14.35 $\pm$ 0.12 & 85.34 $\pm$ 0.17 & 17.05 $\pm$ 0.50 & 16.59 $\pm$ 0.60 & 90.08 $\pm$ 0.38 \\
\midrule
Average & 21.07 $\pm$ 0.25 & 22.27 $\pm$ 0.29 & 85.77 $\pm$ 0.13 & 24.41 $\pm$ 0.23 &	25.14 $\pm$	0.45 &	89.92 $\pm$	0.08 \\
\bottomrule
\end{tabular}
}
\label{tab:stl}
\end{table*}

\begin{table*}[t]
\caption{\small Averaged average precisions for the proposed subsets of Imagenet, with rotation-prediction as the auxiliary task. Each row shows averaged performance across all members of the subset (see Appendix B for complete results). Random detector performance = skew.}
\centering
\small
\scalebox{0.95}{
\begin{tabular}{@{}p{1.1cm}cccccccc@{}}
\toprule
 & & \multicolumn{3}{c}{Classification-only} & \multicolumn{3}{c}{Rotation-augmented} \\
\cmidrule(lr){3-5}
\cmidrule(lr){6-8}
Subset & Skew & MSP & ODIN & Accuracy & MSP & ODIN & Accuracy \\
\midrule
dog & 8.33 & 
23.92 $\pm$ 0.49 & 25.85 $\pm$ 0.09 & 85.09 $\pm$ 0.14 & 24.66 $\pm$ 0.58 & 25.73 $\pm$ 0.87 & 85.25 $\pm$ 0.17 \\
car & 10.00 & 
21.54 $\pm$ 0.62 & 22.49 $\pm$ 0.54 & 77.17 $\pm$ 0.10 & 21.66 $\pm$ 0.19 & 22.38 $\pm$ 0.46 & 76.72 $\pm$ 0.19 \\
snake & 11.11 & 
18.62 $\pm$ 0.93 & 19.18 $\pm$ 0.79 & 69.74 $\pm$ 1.63 & 20.23 $\pm$ 0.18 & 21.17 $\pm$ 0.12 & 70.51 $\pm$ 0.48 \\
spider & 16.67 & 
21.20 $\pm$ 0.56 & 24.15 $\pm$ 0.72 & 68.40 $\pm$ 0.21 & 22.90 $\pm$ 1.29 & 25.10 $\pm$ 1.78 & 68.68 $\pm$ 0.77 \\
fungus & 16.67 & 
42.56 $\pm$ 0.49 & 44.59 $\pm$ 1.46 & 88.23 $\pm$ 0.45 & 44.19 $\pm$ 1.86 & 46.86 $\pm$ 1.13 & 88.47 $\pm$ 0.43
\\
\bottomrule
\end{tabular}
}
\label{tab:imagenet}
\end{table*}

\subsection{Experimental settings}
\textbf{Settings for \cifar and \stl} Our base network for all \cifar experiments is a Wide ResNet~\cite{wrn} with 28 convolutional layers and a widening factor of 10 (WRN-28-10) with the recommended dropout rate of 0.3. Following~\cite{wrn}, we train for 200 epochs, with an initial learning rate of 0.1 which is scaled down by 5 at the 60th, 120th, and 160th epochs, using stochastic gradient descent with Nesterov's momentum at 0.9. We train in parallel on 4 Pascal V100 GPUs with batches of size 128 on each. For \stl, we use the same architecture but append an extra group of 4 residual blocks with the same layer widths as in the previous group. We use a widening factor of 4 instead of 10, and batches of size 64 on each of the 4 GPUs, and train for twice as long. We use the same optimizer settings as with \cifar. In both cases, we apply standard data augmentation of random crops (after padding) and random horizontal reflections. \\ \\
\textbf{Settings for \imagenet} For experiments with the proposed subsets of \imagenet, we replicate the architecture we use for \stl, but add a downsampling average pooling layer after the first convolution on the images. We do not use dropout, and use a batch size of 64, train for 200 epochs; otherwise all other details follow the settings for \stl. The standard data augmentation steps of random crops to a size of $224 \times 224$ and random horizontal reflections are applied. \\ \\
\textbf{Predicting rotation as an auxiliary task} For adding rotation-prediction as an auxiliary task, all we do is append an extra linear layer alongside the one that is responsible for object recognition. $\lambda$ is tuned to 0.5 for \cifar, 1.0 for \stl, and a mix of 0.5 and 1.0 for \imagenet. The optimizer and regularizer settings are kept the same, with the learning rate decayed along with the learning rate for the classifier at the same scales.

We emphasize that this procedure is not equivalent to data augmentation, since we do not optimize the linear classification layer for rotated images. Only the rotation prediction linear layer gets updated for inputs corresponding to the rotation task, and only the linear classification layer gets updated for non-rotated, object-labelled images. Asking the classifier to be rotation-invariant would require the auxiliary task to develop a disjoint subset in the shared representation that is not rotation-invariant, so that it can succeed at predicting rotations. This encourages an internally split representation, thus diminishing the potential advantage we hope to achieve from a shared, mutually beneficial space. \\ \\
\textbf{CPC as an auxiliary task} We also experimented with contrastive predictive coding~\cite{cpc} as an auxiliary task.  Since this is a patch-based method, the input spaces are different across the two tasks:
that of predicting encodings of patches in the image, and that of predicting object category from the
entire image. We found that two tricks are very useful for fostering co-operation: (i) replacing the
normalization layers with their conditional variants~\cite{de2017modulating} (conditioning on the task at hand), and (ii)
using symmetric-padding instead of zero-padding. Since CPC induces significant computational
overhead, we resorted to a lighter-weight base network. This generally comes at the cost of a drop in
classification accuracy and performance at detecting anomalies. We still find, in table 6, that similar
patterns of improvements continue to hold, in terms of improved anomaly detection and improved
generalization, with our auxiliary task. We describe details of the model and report full results in
Appendix C.
\begin{table*}
\caption{\small Averaged average precisions for the proposed subsets of Imagenet where CPC is the auxiliary task. See Appendix C for full results.}
\centering
\small
\scalebox{0.95}{
\begin{tabular}{@{}p{1.1cm}cccccccc@{}}
\toprule
 & & \multicolumn{3}{c}{Classification-only} & \multicolumn{3}{c}{CPC-augmented} \\
\cmidrule(lr){3-5}
\cmidrule(lr){6-8}
Subset & Skew & MSP & ODIN & Accuracy & MSP & ODIN & Accuracy \\
\midrule
dog & 8.33 & 
20.84 $\pm$ 0.50 & 22.77 $\pm$ 0.74 & 83.12 $\pm$ 0.26 & 21.43 $\pm$ 0.63 & 24.08 $\pm$ 0.63 & 84.16 $\pm$ 0.07 \\
car & 10.00 & 
19.86 $\pm$ 0.21 & 21.42 $\pm$ 0.48 & 75.42 $\pm$ 0.11 & 22.21 $\pm$ 0.44 & 23.61 $\pm$ 0.57 & 78.88 $\pm$ 0.15
\\
snake & 11.11 & 
18.20 $\pm$ 0.76 & 18.67 $\pm$ 1.07 & 66.15 $\pm$ 1.89 & 18.78 $\pm$ 0.40 & 20.39 $\pm$ 0.60 & 68.02 $\pm$ 0.85
\\
spider & 16.67 & 
22.03 $\pm$ 0.68 & 24.08 $\pm$ 0.70 & 66.65 $\pm$ 0.42 & 22.28 $\pm$ 0.60 & 23.37 $\pm$ 0.68 & 68.67 $\pm$ 0.36
\\
fungus & 16.67 & 
39.19 $\pm$ 1.26 & 41.71 $\pm$ 1.94 & 87.05 $\pm$ 0.06 & 42.08 $\pm$ 0.57 & 45.05 $\pm$ 1.11 & 88.91 $\pm$ 0.46 \\
\bottomrule
\end{tabular}
}
\label{tab:cpc-imagenet}
\end{table*}

\subsection{Discussion}
\textbf{Self-supervised multi-task learning is effective} In Tables~\ref{tab:cifar} and~\ref{tab:stl} we report average precision scores on \cifar and \stl for the baseline scoring methods MSP~\cite{danbaseline} and ODIN~\cite{odin}. We note that ODIN, with fixed hyperparameter settings across all experiments, continues to outperform MSP most of the time. When we augment our classifiers with the auxiliary rotation-prediction task, we find that anomaly detection as well as test set accuracy are markedly improved for both scoring methods. As we have remarked earlier, a representation space with greater semanticity should be expected to bring improvements on both fronts. All results report mean $\pm$ standard deviation over 3 trials. In Table~\ref{tab:imagenet}, we repeat the same process for the much harder Imagenet subsets. Full results, corresponding to individual members of the subsets, are in Appendix B, while here we only show the average performance across all members of the subset. In Appendix C, we show results when CPC is the auxiliary task. Taken together, our results indicate that multi-task learning with self-supervised auxiliary tasks can be an effective approach for improving anomaly detection, with accompanying improvements in generalization. \\ \\
\textbf{Improved test set accuracy is not enough} Training methods developed solely to improve generalization, without consideration of the affect on semantic understanding, might perform worse at detecting semantic anomalies. This is because it is often possible to pick up on low-level or contextual discriminatory patterns, which are almost surely biased in relatively small datasets for complex domains such as natural images, and perform reasonably well on the test set. To illustrate this, we run an experiment where we randomly mask out a $16 \times 16$ region in \cifar images from within the central $21 \times 21$ region. We see below that while this leads to improved test accuracies, anomaly detection suffers (numbers are averages across hold-out-class trials). 
\begin{center}
\small
\begin{tabular}{@{}p{3.3cm}cc@{}}
    \toprule
    Method & Accuracy & Av. Prec. with MSP \\
    \midrule
    Base model & 96.03 $\pm$ 0.00 & 34.92 $\pm$ 0.41 \\
    Random-center-masked & 96.27 $\pm$ 0.05 & 34.41 $\pm$ 0.74 \\
    Rotation-augmented & 96.83 $\pm$ 0.02 & 40.22 $\pm$ 0.16 \\
    \bottomrule
\end{tabular}
\end{center}
This suggests that while the masking strategy is effective as a regularizer, it might come at the cost of less semantic representation. Certain training choices can therefore result in models with seemingly improved generalization but which have poorer representation for tasks that require a more coherent understanding. For comparison, the rotation-augmented network achieves both a higher test set accuracy as well as an improved average precision. This example serves as a caution toward developing techniques that might achieve reassuring test set performance, while inadvertently following an internal \textit{modus operandi} that is misaligned with the pattern of reasoning we hope they discover. This can have unexpected consequences when models trained with such methods are deployed in the real world.

\section{Conclusion}
We provided a critical review of the current interest in OOD detection, concluding that realistic applications involve detecting semantic distributional shift for a specified context, which we regard as anomaly detection. While there is significant recent interest in the area, current research suffers from questionable benchmarks and methodology. In light of these considerations, we suggested a set of benchmarks which are better aligned with realistic anomaly detection applications in the context of object classification systems.

We also explored the effectiveness of a multi-task learning framework with auxiliary objectives. Our results demonstrate improved anomaly detection along with improved generalization under such augmented objectives. This suggests that inductive biases induced through such auxiliary tasks could have an important role to play in developing more trustworthy neural networks. 

We note that the ability to detect semantic anomalies also provides us with an indirect view of semanticity in the representations learned by our mostly opaque deep models.

\section*{Acknowledgements}
We thank Rachel Rolland for referencing and discussing the motivating examples of anomaly
detection in nature studies. Ishaan Gulrajani, Tim Cooijmans, and anonymous reviewers provided
useful feedback.

This work was enabled by the computational resources provided by Compute Canada and funding support from the Canadian CIFAR AI chair and NSERC Discovery Grant. 

\Urlmuskip=0mu plus 2mu\relax
\bibliographystyle{unsrt}
\bibliography{references}

\newpage
\appendix
\section{Imagenet benchmarks}
We present details of the Imagenet-based benchmark we proposed. For constructing these datasets, we first sorted all subsets by the number of members, as structured in the Imagenet hierarchy. We then picked from among the list of top twenty subsets, with a preference for subsets that are more closely aligned with the theme of motivating practical applications we provided. We also manually inspected the data, to check for inconsistencies, and performed some pruning. For example, the \emph{beetle} subset, while seeming ideal, has some issues with labelling: \emph{leaf beetle} and \emph{ladybug} appear to overlap in some cases. Finally, we settled on our choice of 5 subsets.  In table~\ref{tab:breakdown}, we list the members under every proposed subset.
\begin{table*}[h]
\caption{Imagenet subset members}
\centering
\begin{tabular}{ccccc}
\toprule
Dog (hound) & Car & Snake (colubrid) & Spider & Fungus \\
\midrule
Ibizan hound & Model T & ringneck snake & tarantula & stinkhorn \\
bluetick & race car & vine snake & Argiope aurantia & bolete \\
beagle & sports car & hognose snake & barn spider & hen-of-the-woods \\
Afghan hound & minivan & thunder snake & black widow & earthstar \\
Weimaraner & ambulance & garter snake & garden spider & gyromitra \\
Saluki & cab & king snake & wolf spider & coral fungus \\
redbone & beach wagon & night snake & & \\
otterhound & jeep & green snake & & \\
Norweigian elkhound & convertible & water snake & & \\
basset hound & limo & & & \\
Scottish deerhound & & & & \\
bloodhound & & & & \\
\bottomrule
\end{tabular}
\label{tab:breakdown}
\end{table*}

\begin{figure}
\centering
     \begin{subfigure}[t]{0.25\linewidth}
     \centering
         \includegraphics[scale=0.1]{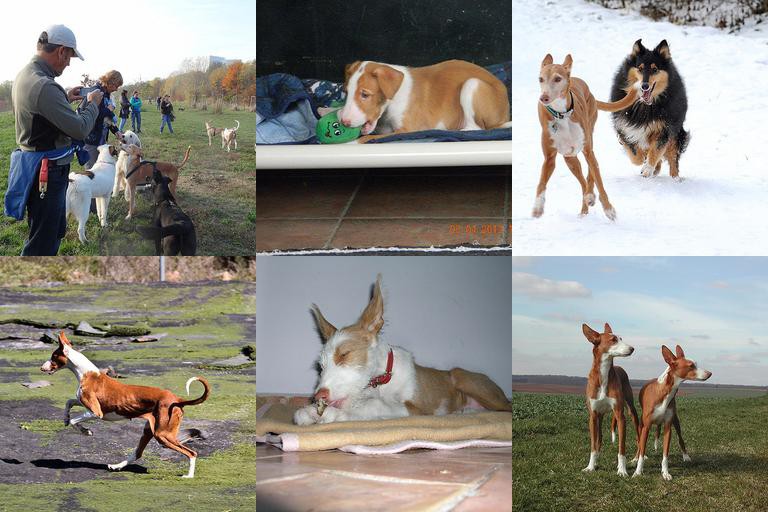}
         \caption{Ibizan hound}
     \end{subfigure}%
     \hfill
     \begin{subfigure}[t]{0.25\linewidth}
         \centering
         \includegraphics[scale=0.1]{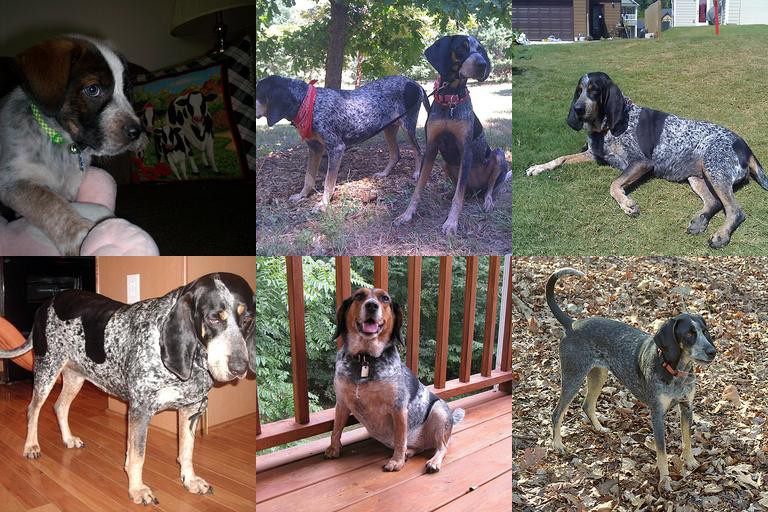}
         \caption{Bluetick}
     \end{subfigure}%
     \hfill
     \begin{subfigure}[t]{0.25\linewidth}
         \centering
         \includegraphics[scale=0.1]{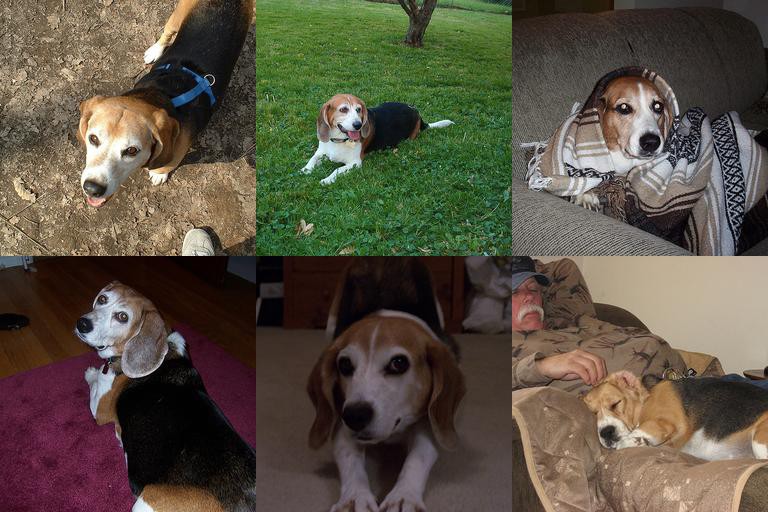}
         \caption{Beagle}
     \end{subfigure}
     \\
     \begin{subfigure}[t]{0.25\linewidth}
     \centering
         \includegraphics[scale=0.1]{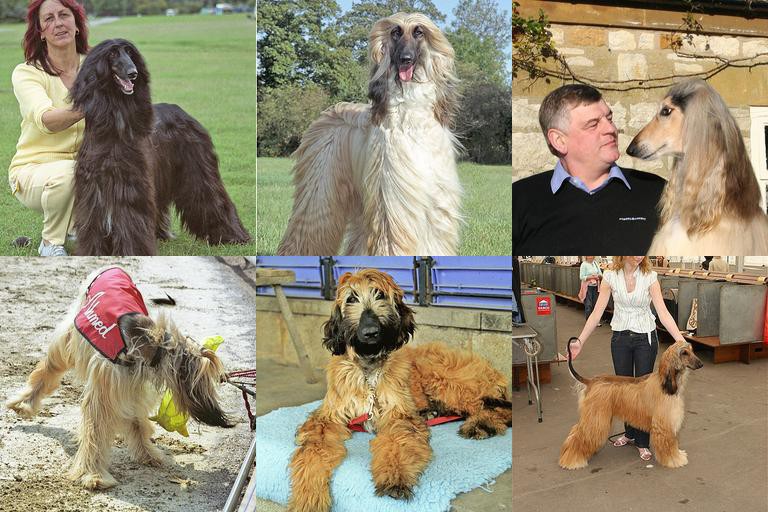}
         \caption{Afghan hound}
     \end{subfigure}%
     \hfill
     \begin{subfigure}[t]{0.25\linewidth}
         \centering
         \includegraphics[scale=0.1]{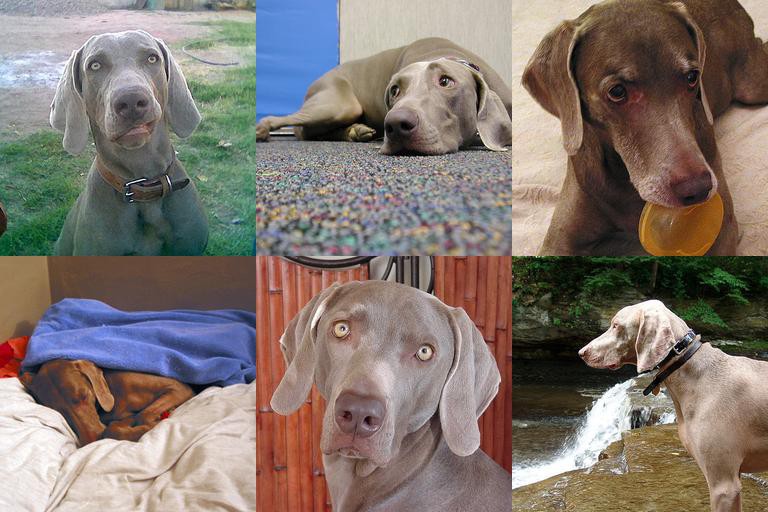}
         \caption{Weimaraner}
     \end{subfigure}%
     \hfill
     \begin{subfigure}[t]{0.25\linewidth}
         \centering
         \includegraphics[scale=0.1]{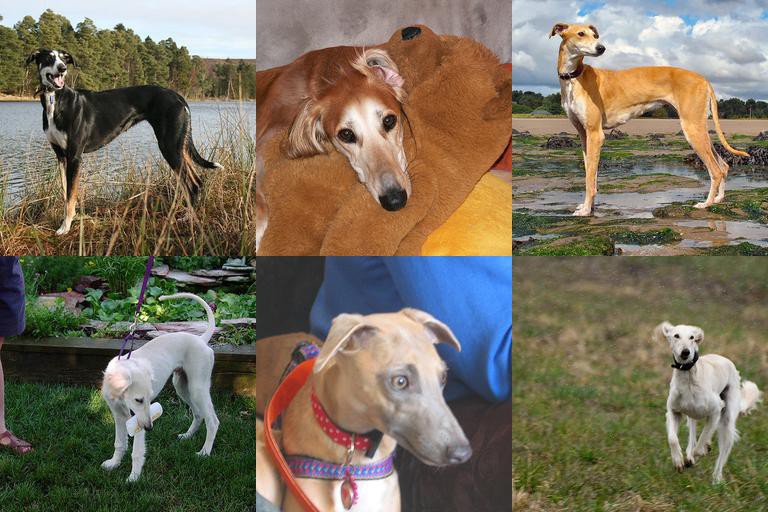}
         \caption{Saluki}
     \end{subfigure}
     \\
     \begin{subfigure}[t]{0.25\linewidth}
     \centering
         \includegraphics[scale=0.1]{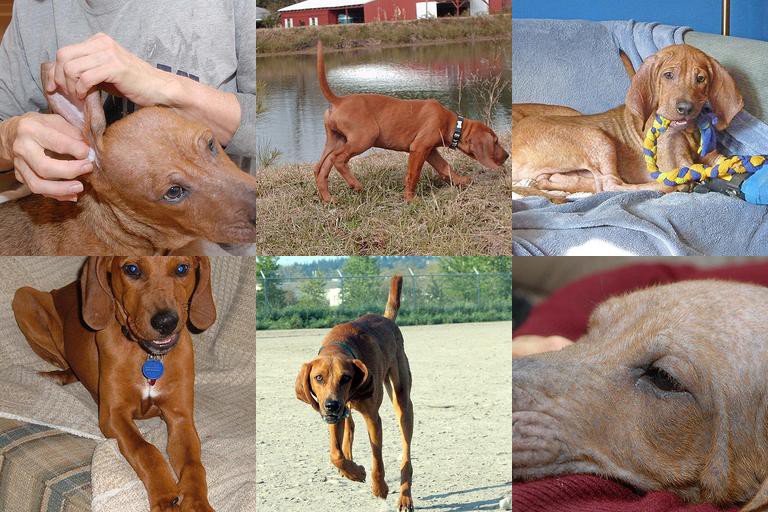}
         \caption{Redbone}
     \end{subfigure}%
     \hfill
     \begin{subfigure}[t]{0.25\linewidth}
         \centering
         \includegraphics[scale=0.1]{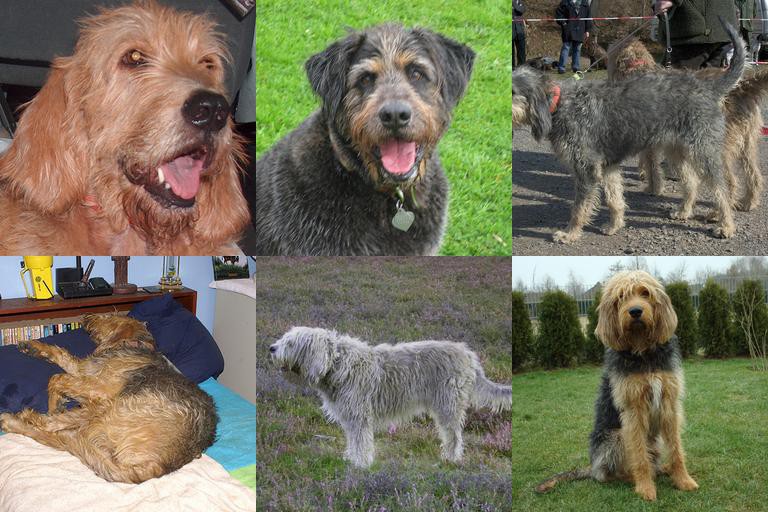}
         \caption{Otterhound}
     \end{subfigure}%
     \hfill
     \begin{subfigure}[t]{0.25\linewidth}
         \centering
         \includegraphics[scale=0.1]{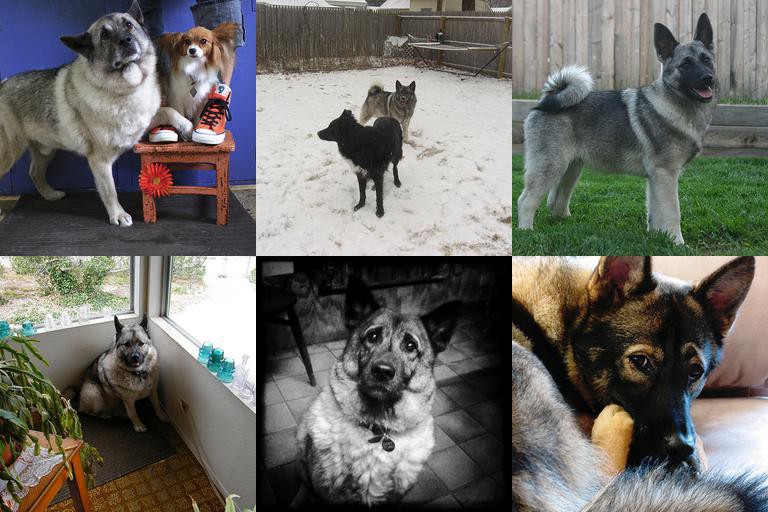}
         \caption{Norweigian elkhound}
     \end{subfigure}
     \\
     \begin{subfigure}[t]{0.25\linewidth}
     \centering
         \includegraphics[scale=0.1]{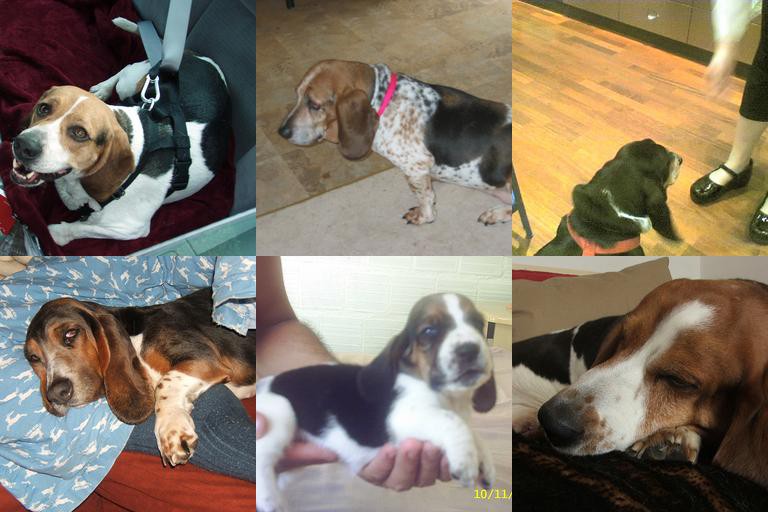}
         \caption{Basset hound}
     \end{subfigure}%
     \hfill
     \begin{subfigure}[t]{0.25\linewidth}
         \centering
         \includegraphics[scale=0.1]{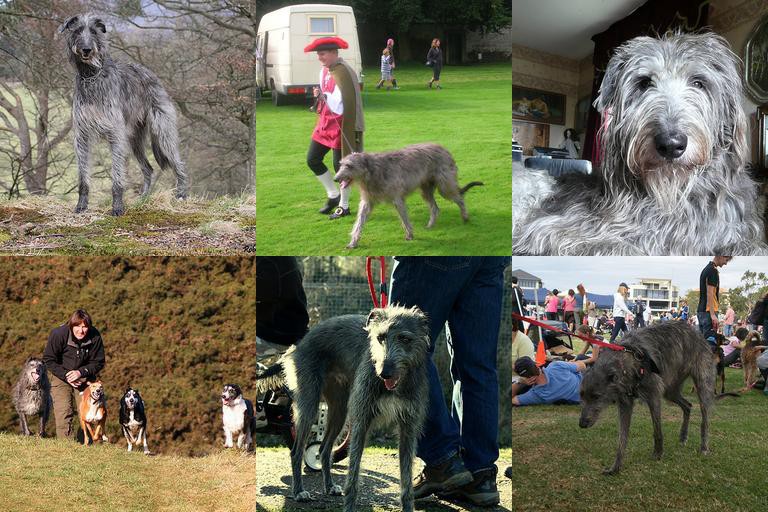}
         \caption{Scottish deerhound}
     \end{subfigure}%
     \hfill
     \begin{subfigure}[t]{0.25\linewidth}
         \centering
         \includegraphics[scale=0.1]{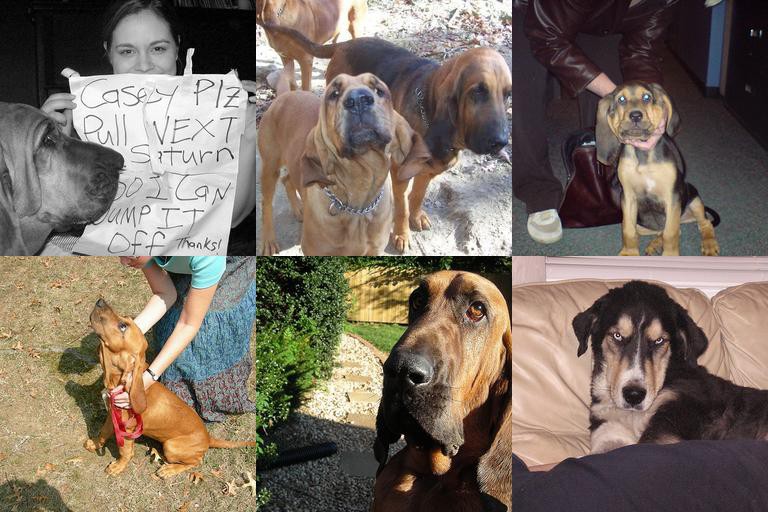}
         \caption{Bloodhound}
     \end{subfigure}
        \caption{Dog (hound dog)}
        \label{hounddogs}
\end{figure}

\begin{figure}
\centering
     \begin{subfigure}[t]{0.25\linewidth}
     \centering
         \includegraphics[scale=0.1]{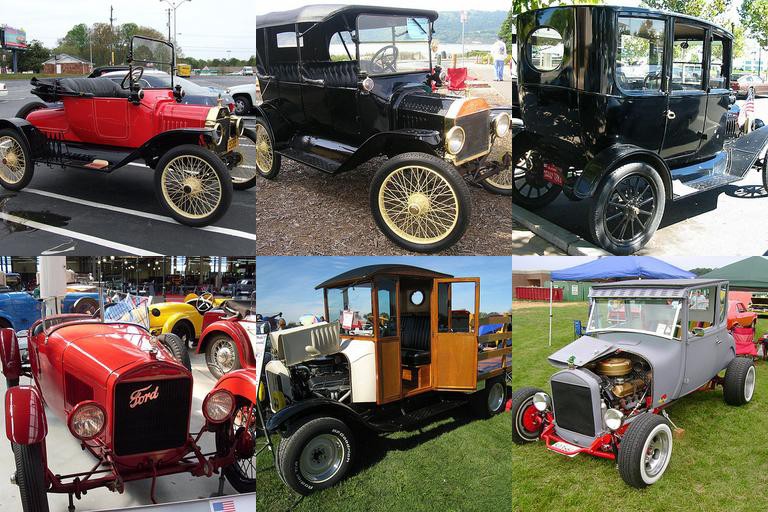}
         \caption{Model T}
     \end{subfigure}%
     \hfill
     \begin{subfigure}[t]{0.25\linewidth}
         \centering
         \includegraphics[scale=0.1]{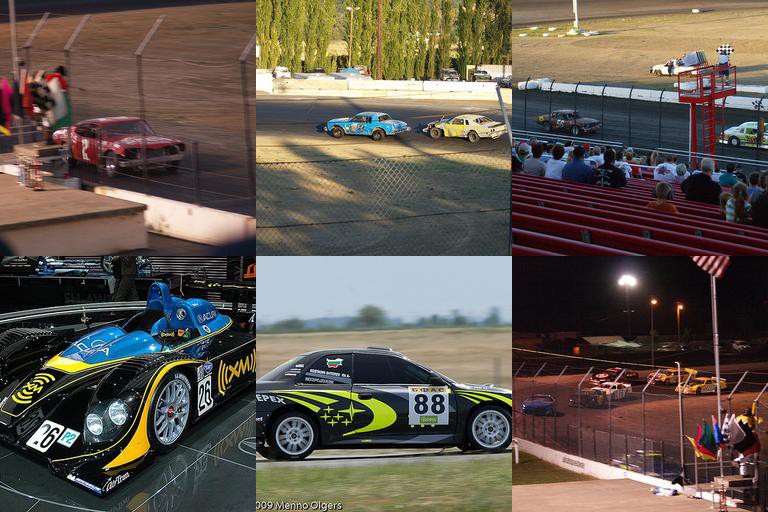}
         \caption{Race car}
     \end{subfigure}%
     \hfill
     \begin{subfigure}[t]{0.25\linewidth}
         \centering
         \includegraphics[scale=0.1]{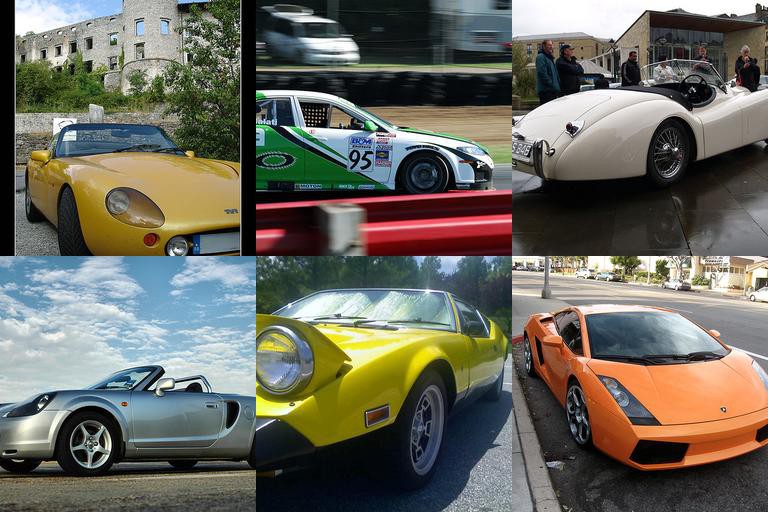}
         \caption{Sports car}
     \end{subfigure}
     \\
     \begin{subfigure}[t]{0.25\linewidth}
     \centering
         \includegraphics[scale=0.1]{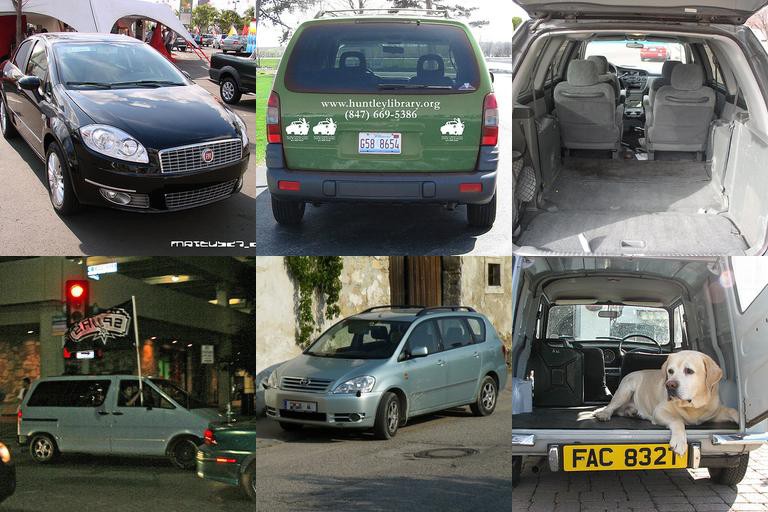}
         \caption{Minivan}
     \end{subfigure}%
     \hfill
     \begin{subfigure}[t]{0.25\linewidth}
         \centering
         \includegraphics[scale=0.1]{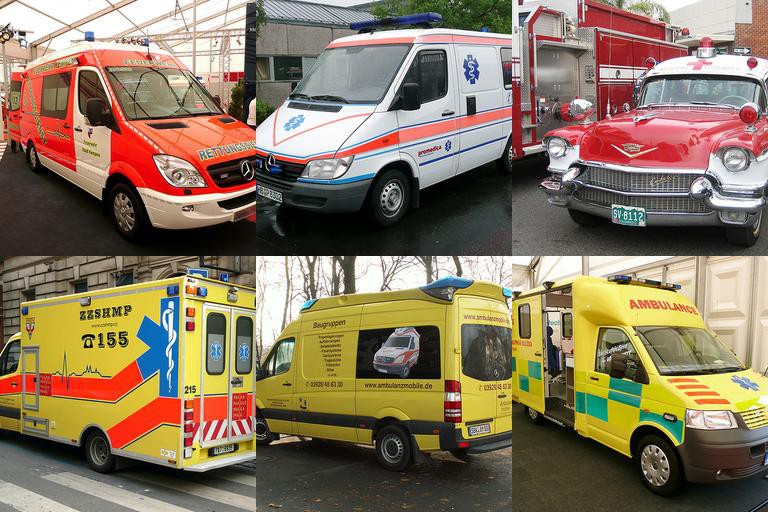}
         \caption{Ambulance}
     \end{subfigure}%
     \hfill
     \begin{subfigure}[t]{0.25\linewidth}
         \centering
         \includegraphics[scale=0.1]{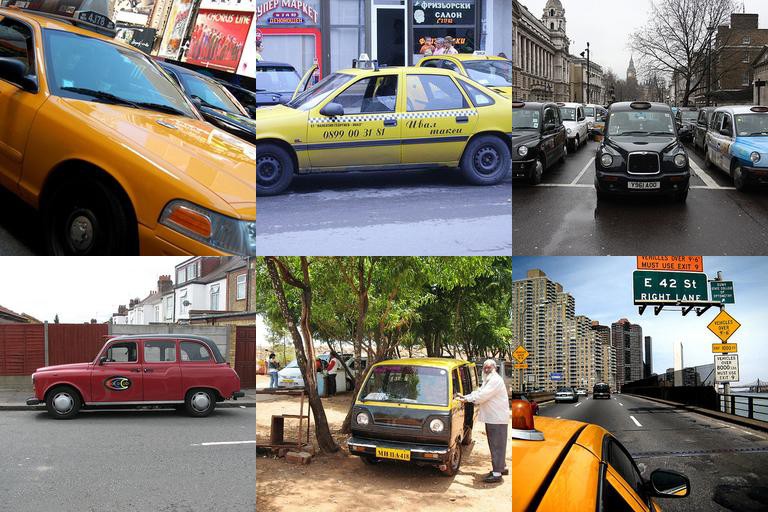}
         \caption{Cab}
     \end{subfigure}
     \\
     \begin{subfigure}[t]{0.25\linewidth}
     \centering
         \includegraphics[scale=0.1]{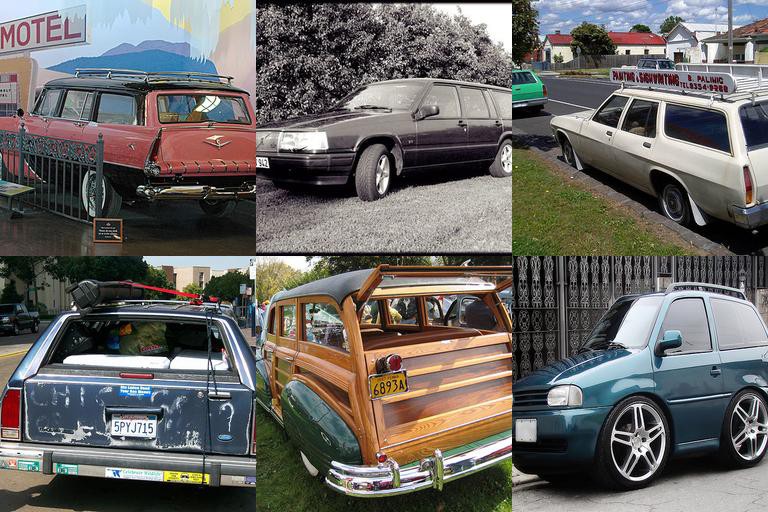}
         \caption{Beach wagon}
     \end{subfigure}%
     \hfill
     \begin{subfigure}[t]{0.25\linewidth}
         \centering
         \includegraphics[scale=0.1]{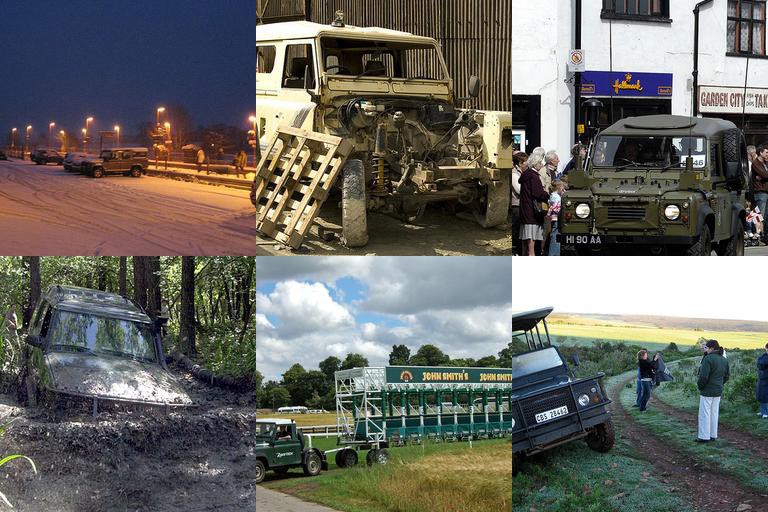}
         \caption{Jeep}
     \end{subfigure}%
     \hfill
     \begin{subfigure}[t]{0.25\linewidth}
         \centering
         \includegraphics[scale=0.1]{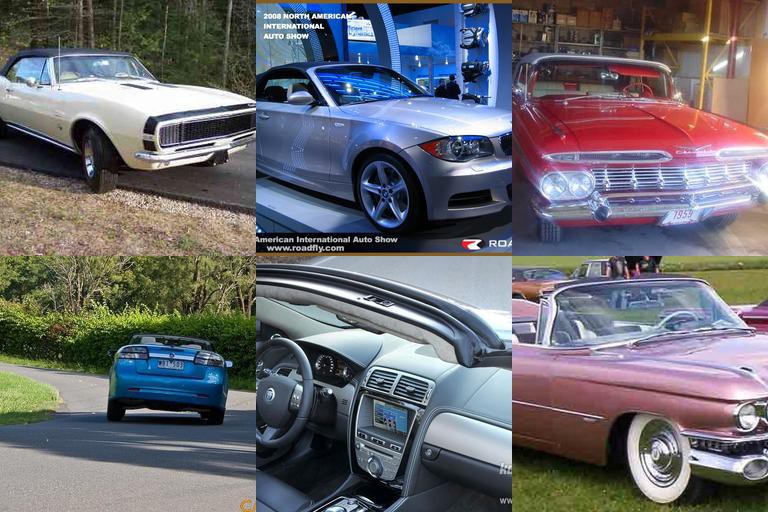}
         \caption{Convertile}
     \end{subfigure}
     \\
     \begin{subfigure}[t]{0.25\linewidth}
     \centering
         \includegraphics[scale=0.1]{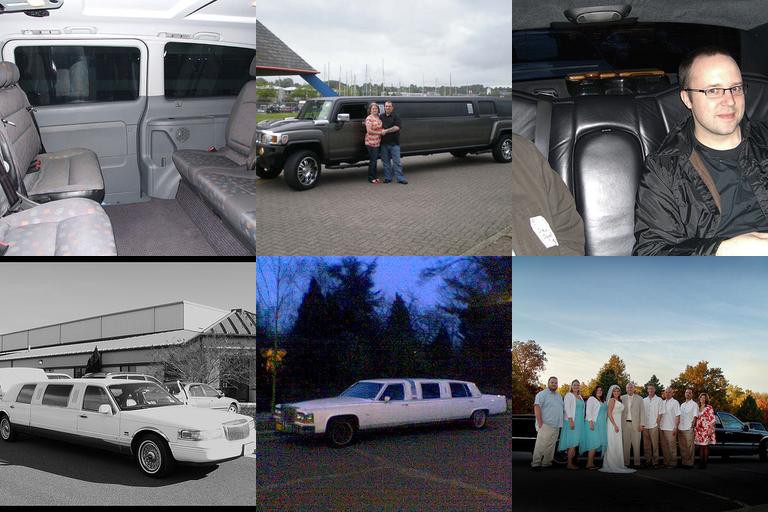}
         \caption{Limousine}
     \end{subfigure}%
        \caption{Car}
        \label{fig:cars}
\end{figure}

\begin{figure}
\centering
     \begin{subfigure}[t]{0.25\linewidth}
     \centering
         \includegraphics[scale=0.1]{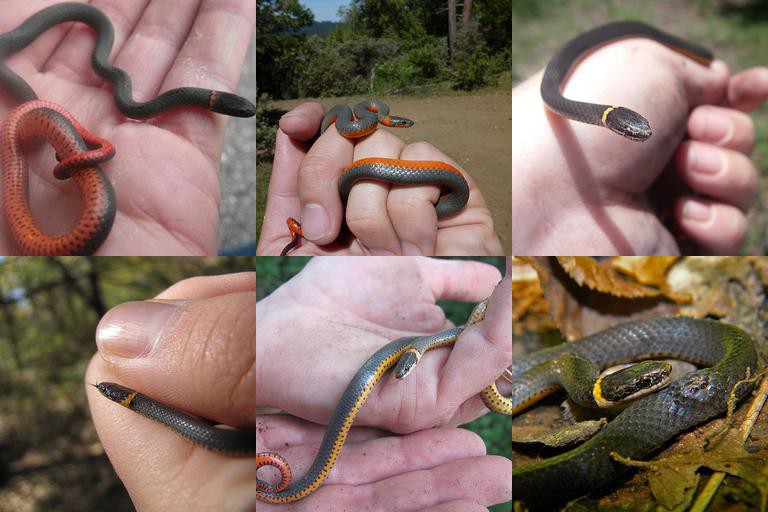}
         \caption{Ringneck snake}
     \end{subfigure}%
     \hfill
     \begin{subfigure}[t]{0.25\linewidth}
         \centering
         \includegraphics[scale=0.1]{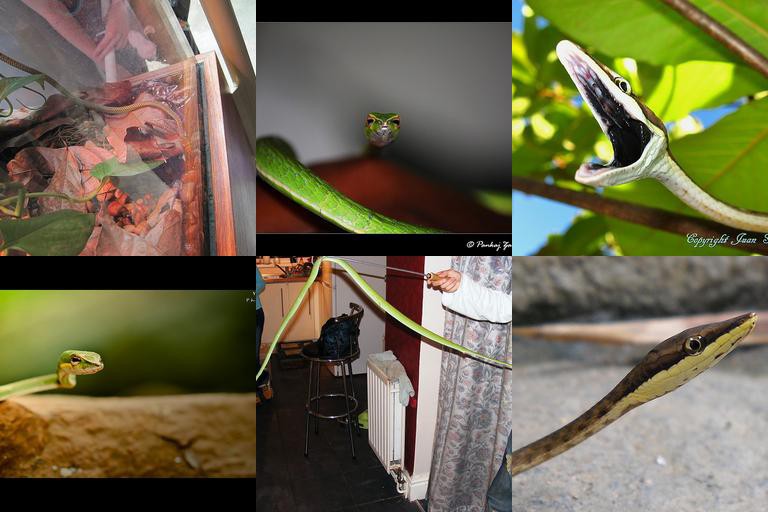}
         \caption{Vine snake}
     \end{subfigure}%
     \hfill
     \begin{subfigure}[t]{0.25\linewidth}
         \centering
         \includegraphics[scale=0.1]{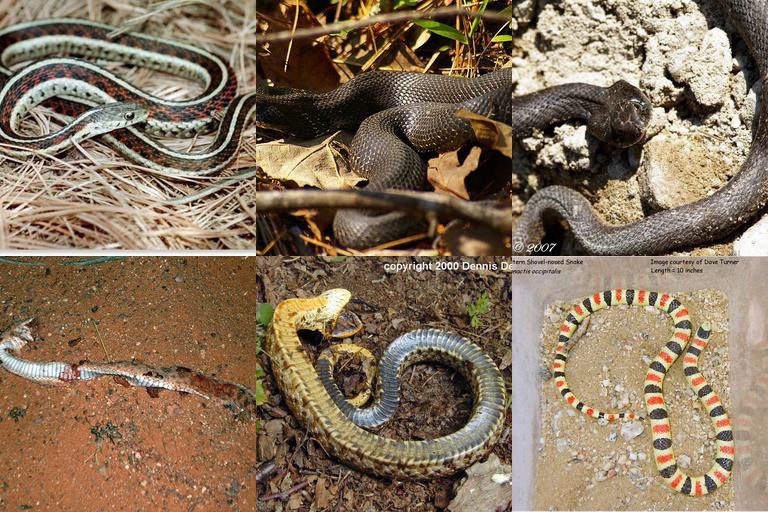}
         \caption{Hognose snake}
     \end{subfigure}
     \\
     \begin{subfigure}[t]{0.25\linewidth}
     \centering
         \includegraphics[scale=0.1]{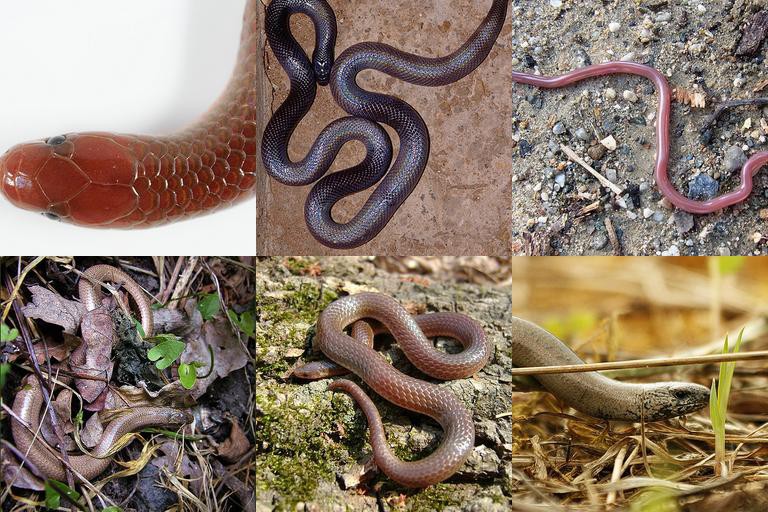}
         \caption{Thunder snake}
     \end{subfigure}%
     \hfill
     \begin{subfigure}[t]{0.25\linewidth}
         \centering
         \includegraphics[scale=0.1]{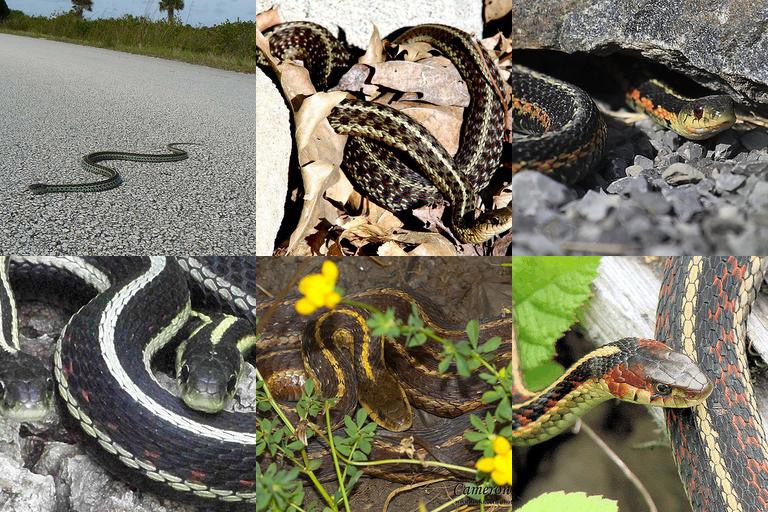}
         \caption{Garter snake}
     \end{subfigure}%
     \hfill
     \begin{subfigure}[t]{0.25\linewidth}
         \centering
         \includegraphics[scale=0.1]{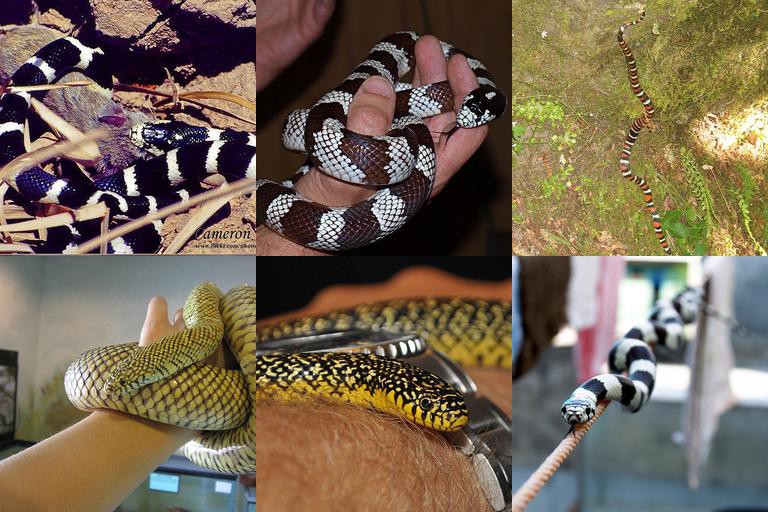}
         \caption{King snake}
     \end{subfigure}
     \\
     \begin{subfigure}[t]{0.25\linewidth}
     \centering
         \includegraphics[scale=0.1]{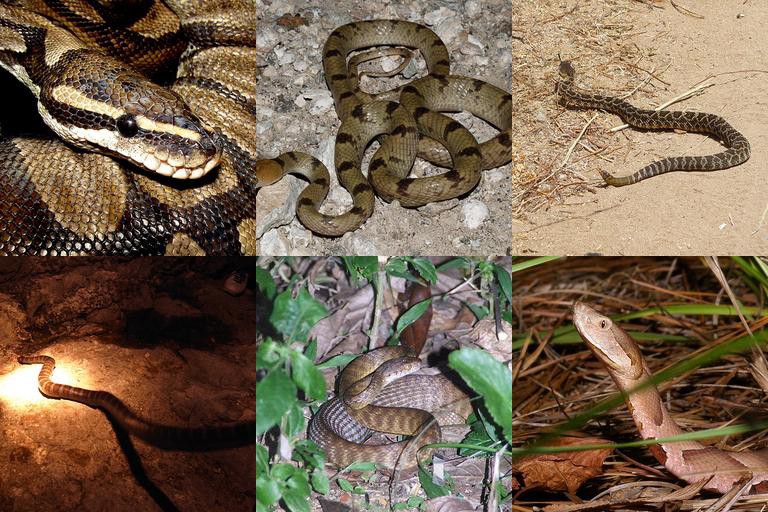}
         \caption{Night snake}
     \end{subfigure}%
     \hfill
     \begin{subfigure}[t]{0.25\linewidth}
         \centering
         \includegraphics[scale=0.1]{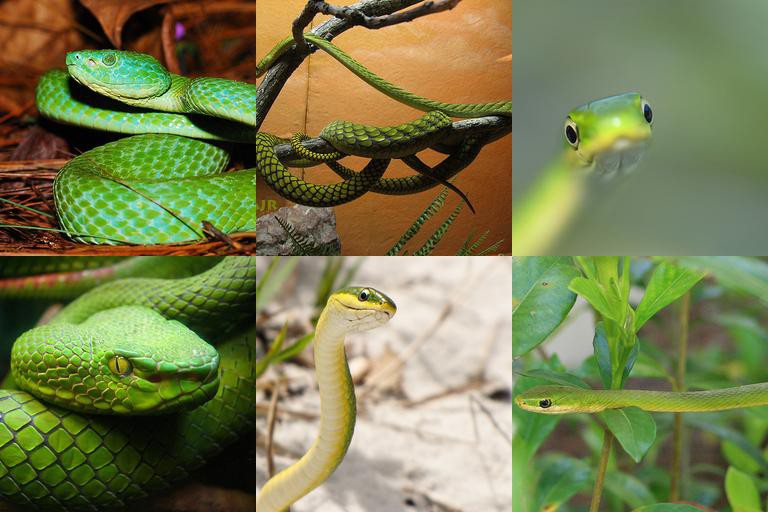}
         \caption{Green snake}
     \end{subfigure}%
     \hfill
     \begin{subfigure}[t]{0.25\linewidth}
         \centering
         \includegraphics[scale=0.1]{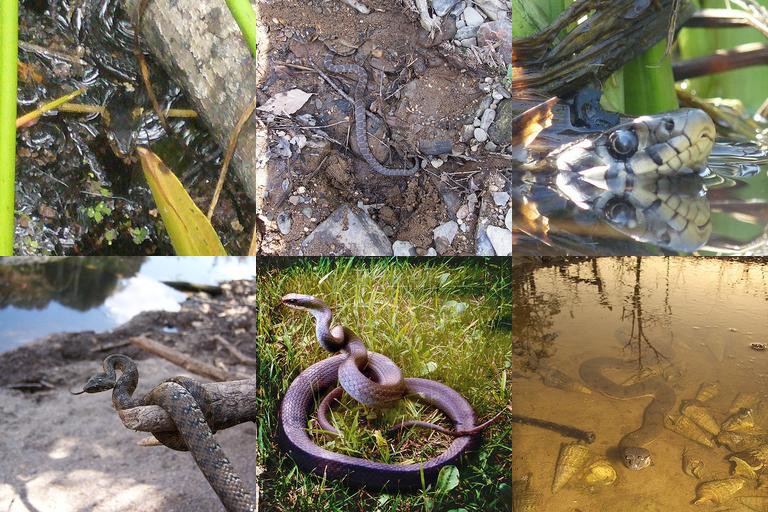}
         \caption{Water snake}
     \end{subfigure}
     \caption{Snake (colubrid)}
     \label{fig:snakes}
\end{figure}

\begin{figure}
\centering
     \begin{subfigure}[t]{0.25\linewidth}
     \centering
         \includegraphics[scale=0.1]{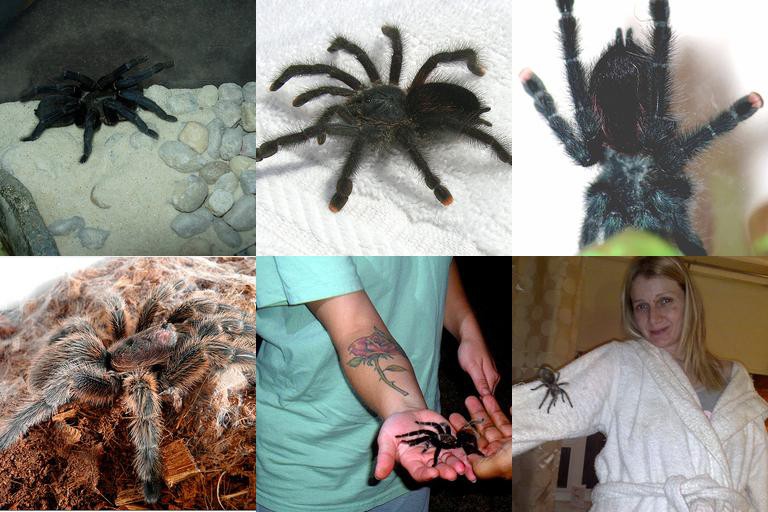}
         \caption{Tarantula}
     \end{subfigure}%
     \hfill
     \begin{subfigure}[t]{0.25\linewidth}
         \centering
         \includegraphics[scale=0.1]{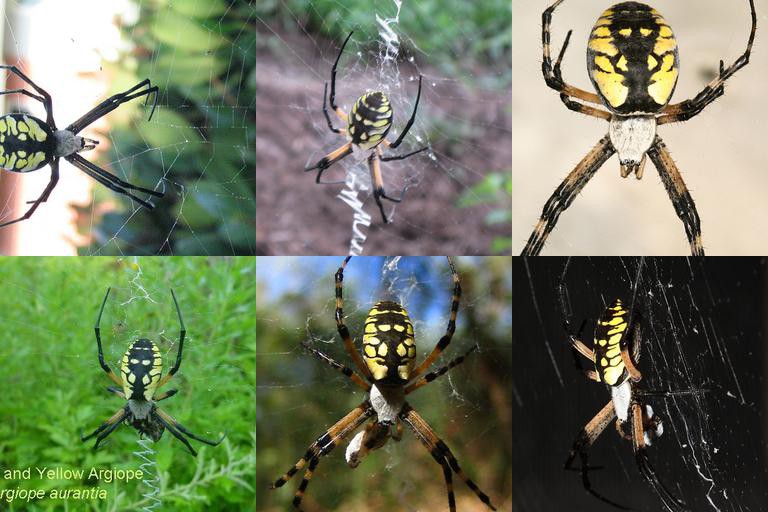}
         \caption{Argiope aurantia}
     \end{subfigure}%
     \hfill
     \begin{subfigure}[t]{0.25\linewidth}
         \centering
         \includegraphics[scale=0.1]{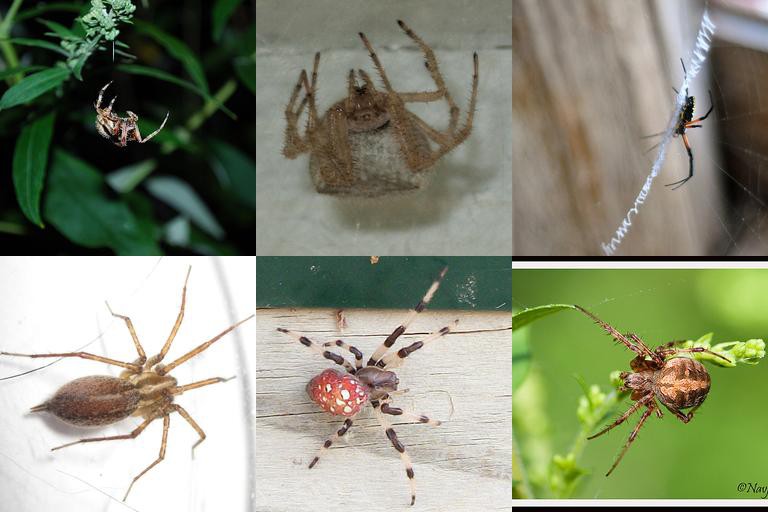}
         \caption{Barn spider}
     \end{subfigure}
     \\
     \begin{subfigure}[t]{0.25\linewidth}
     \centering
         \includegraphics[scale=0.1]{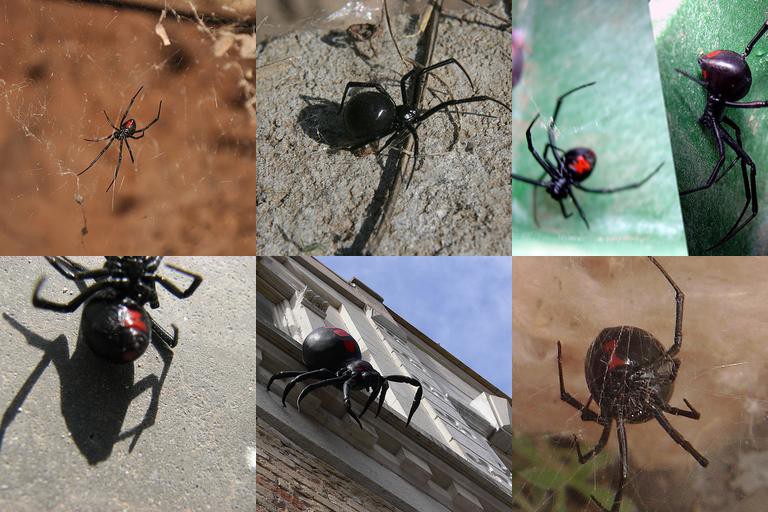}
         \caption{Black widow}
     \end{subfigure}%
     \hfill
     \begin{subfigure}[t]{0.25\linewidth}
         \centering
         \includegraphics[scale=0.1]{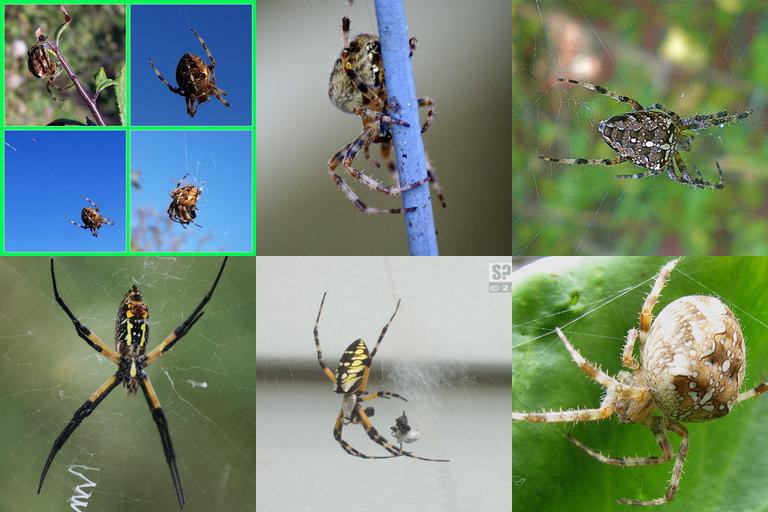}
         \caption{Garden spider}
     \end{subfigure}%
     \hfill
     \begin{subfigure}[t]{0.25\linewidth}
         \centering
         \includegraphics[scale=0.1]{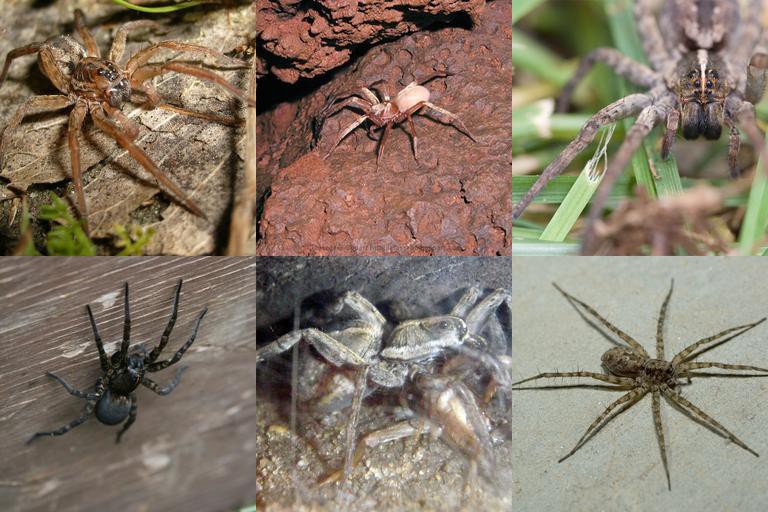}
         \caption{Wolf spider}
     \end{subfigure}
     \caption{Spider}
     \label{fig:spiders}
\end{figure}

\begin{figure}
\centering
     \begin{subfigure}[t]{0.25\linewidth}
     \centering
         \includegraphics[scale=0.1]{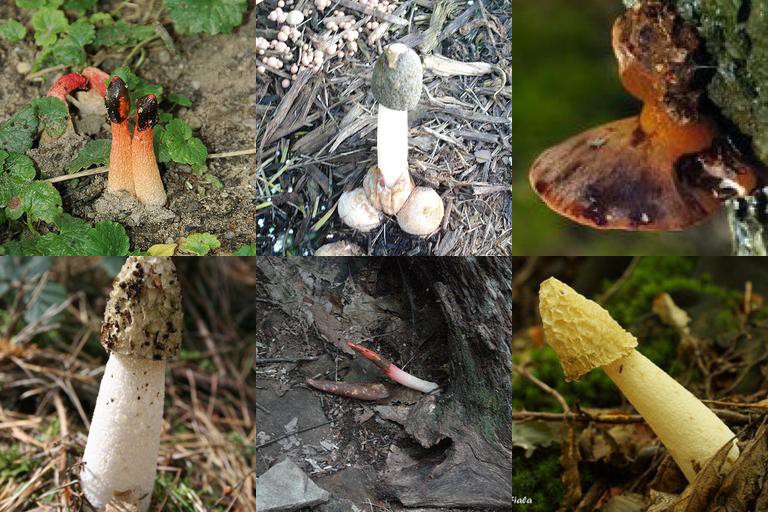}
         \caption{Stinkhorn}
     \end{subfigure}%
     \hfill
     \begin{subfigure}[t]{0.25\linewidth}
         \centering
         \includegraphics[scale=0.1]{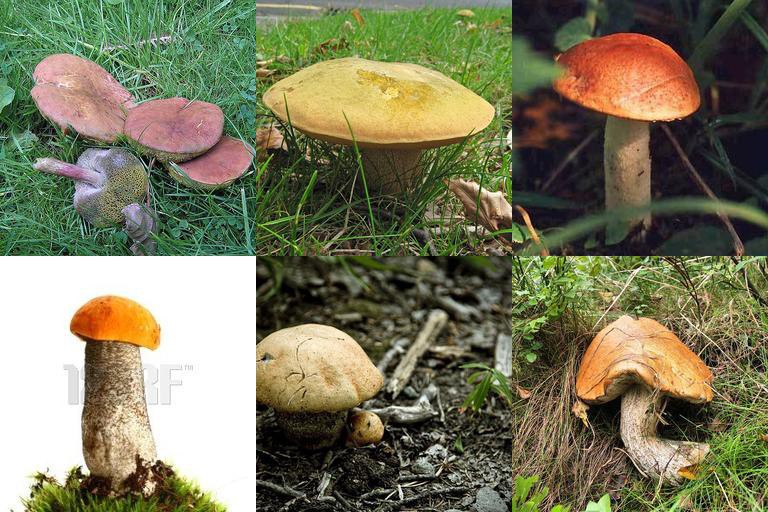}
         \caption{Bolete}
     \end{subfigure}%
     \hfill
     \begin{subfigure}[t]{0.25\linewidth}
         \centering
         \includegraphics[scale=0.1]{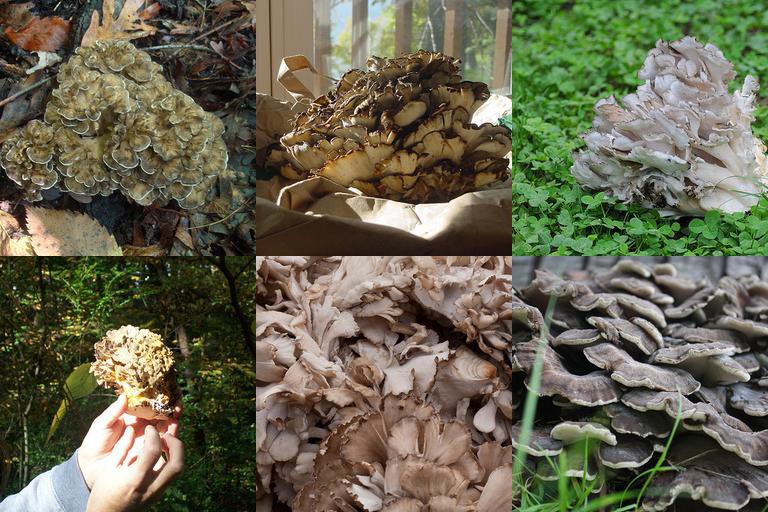}
         \caption{Hen-of-the-woods}
     \end{subfigure}
     \\
     \begin{subfigure}[t]{0.25\linewidth}
     \centering
         \includegraphics[scale=0.1]{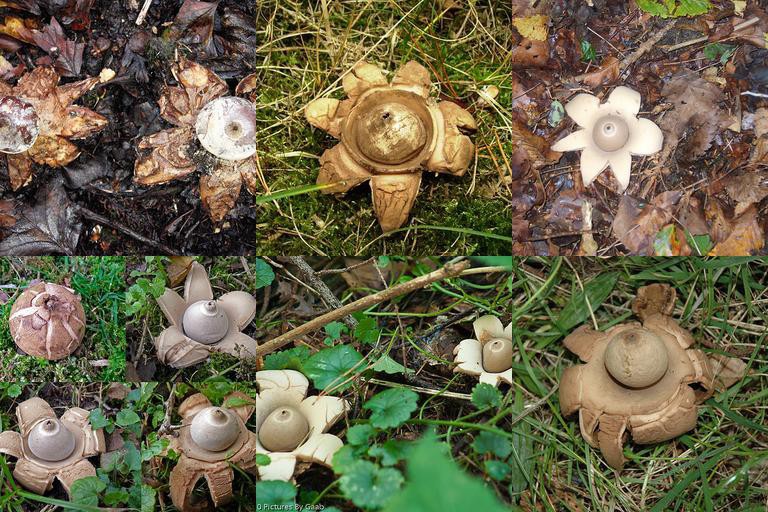}
         \caption{Earthstar}
     \end{subfigure}%
     \hfill
     \begin{subfigure}[t]{0.25\linewidth}
         \centering
         \includegraphics[scale=0.1]{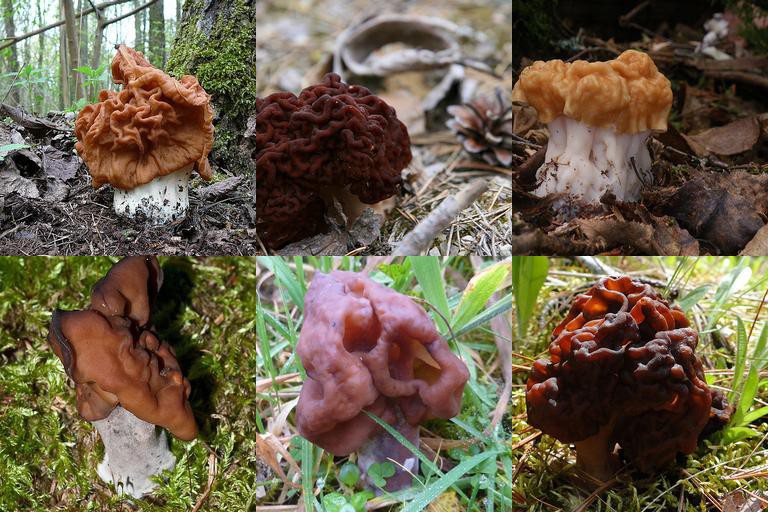}
         \caption{Gyromitra}
     \end{subfigure}%
     \hfill
     \begin{subfigure}[t]{0.25\linewidth}
         \centering
         \includegraphics[scale=0.1]{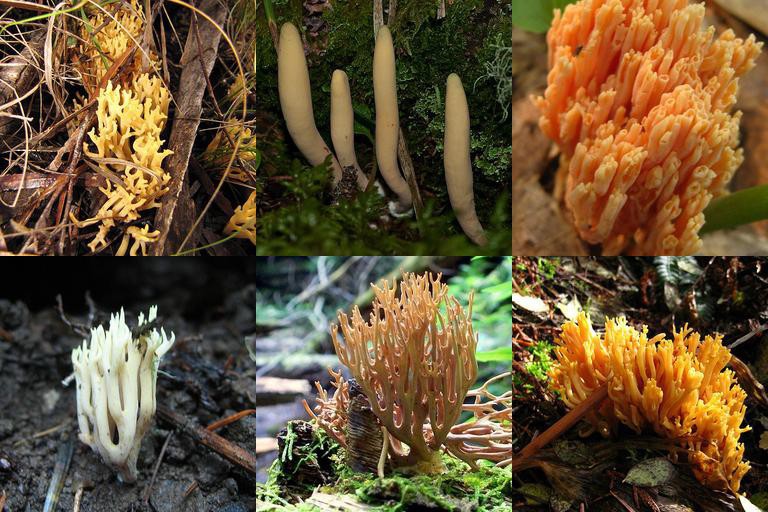}
         \caption{Coral fungus}
     \end{subfigure}
     \caption{Fungus}
     \label{fig:fungi}
\end{figure}

\noindent In Figures~\ref{hounddogs},\ref{fig:cars},\ref{fig:snakes},\ref{fig:spiders},\ref{fig:fungi} we show samples of images. The sets are collected by first resizing such that the shorter side is of length 256 pixels, followed by a center crop. It is obvious that due to intrinsic dataset bias, some categories may be viewed as anomalous without careful inspection of the object of interest. For example, owing to their smaller size, ringneck snakes are most often photographed when held in human hands, and race cars are usually pictured on race tracks. Such dataset biases have historically been hard to account for, and we recommend more thoughtful curation of specific datasets for specific tasks for our proposed style of benchmarks to be more reflective of field performance. This is also why we recommend multiple hold-out trials as opposed to a single predefined split: it is possible that a particular set of such biases fall into a particular split, which would score methods that exploit such biases higher than ones that are potentially more robust across a family of biases. Multiple hold-out-trials reduce such inadvertent advantages to bias.

\clearpage

\section{Expanded results for Imagenet-subset experiments with rotation-prediction as the auxiliary task}
We show expanded results for the Imagenet experiments with predicting rotation as an auxiliary task, corresponding to every hold-out-class experiment, in the tables below.
\begin{center}
\centering
\scriptsize
\begin{tabular}{p{3.0cm}cccccccc}
\toprule
 & \multicolumn{3}{c}{Classification-only} & \multicolumn{3}{c}{Rotation-augmented} \\
\cmidrule(lr){2-4}
\cmidrule(lr){5-7}
Anomalous dogs & MSP & ODIN & Accuracy & MSP & ODIN & Accuracy \\
\midrule
Ibizan hound & 25.56 $\pm$ 2.43 & 26.34 $\pm$ 2.65 & 85.19 $\pm$ 0.65 & 22.85 $\pm$ 2.54 & 24.27 $\pm$ 2.60 & 85.99 $\pm$ 0.73 \\
bluetick & 34.37 $\pm$ 4.32 & 39.19 $\pm$ 1.43 & 85.50 $\pm$ 0.82 & 29.60 $\pm$ 4.39 & 31.70 $\pm$ 1.99 & 85.36 $\pm$ 0.59 \\
beagle! & 18.29 $\pm$ 1.54 & 17.05 $\pm$ 1.21 & 86.33 $\pm$ 1.18 & 19.79 $\pm$ 2.02 & 18.52 $\pm$ 1.69 & 86.37 $\pm$ 0.71 \\
Afghan hound & 20.05 $\pm$ 3.07 & 18.69 $\pm$ 1.22 & 84.16 $\pm$ 0.59 & 20.62 $\pm$ 1.26 & 20.56 $\pm$ 1.37 & 83.44 $\pm$ 0.98 \\
Weimaraner & 31.04 $\pm$ 2.22 & 36.87 $\pm$ 2.90 & 83.68 $\pm$ 0.36 & 27.65 $\pm$ 2.52 & 30.59 $\pm$ 1.03 & 83.62 $\pm$ 1.40 \\
Saluki & 26.64 $\pm$ 2.50 & 31.75 $\pm$ 2.01 & 83.76 $\pm$ 1.08 & 28.20 $\pm$ 1.46 & 29.27 $\pm$ 1.30 & 84.74 $\pm$ 0.19 \\
redbone & 17.93 $\pm$ 0.59 & 18.66 $\pm$ 0.59 & 86.61 $\pm$ 1.48 & 19.14 $\pm$ 0.39 & 19.54 $\pm$ 1.15 & 86.01 $\pm$ 1.10 \\
otterhound & 22.71 $\pm$ 0.77 & 23.31 $\pm$ 0.83 & 84.50 $\pm$ 0.54 & 21.32 $\pm$ 3.71 & 22.90 $\pm$ 4.28 & 84.24 $\pm$ 0.56 \\
Norweigian elkhound & 28.82 $\pm$ 2.16 & 36.55 $\pm$ 0.61 & 83.91 $\pm$ 1.35 & 34.64 $\pm$ 6.16 & 41.61 $\pm$ 6.55 & 84.33 $\pm$ 0.84 \\
basset hound & 18.39 $\pm$ 0.91 & 16.23 $\pm$ 0.58 & 86.34 $\pm$ 0.76 & 21.33 $\pm$ 3.10 & 19.46 $\pm$ 1.72 & 86.45 $\pm$ 0.65 \\
Scottish deerhound & 26.83 $\pm$ 2.97 & 26.52 $\pm$ 2.23 & 83.95 $\pm$ 0.61 & 26.64 $\pm$ 1.17 & 24.87 $\pm$ 0.52 & 83.80 $\pm$ 0.02 \\
bloodhound & 16.43 $\pm$ 1.17 & 19.04 $\pm$ 0.91 & 87.17 $\pm$ 0.24 & 24.19 $\pm$ 6.23 & 25.42 $\pm$ 6.60 & 88.69 $\pm$ 0.84 \\
\midrule
Average & 23.92 $\pm$ 0.49 & 25.85 $\pm$ 0.09 & 85.09 $\pm$ 0.14 & 24.66 $\pm$ 0.58 & 25.73 $\pm$ 0.87 & 85.25 $\pm$ 0.17 \\
\bottomrule
\end{tabular}
\end{center}

\begin{center}
\centering
\scriptsize
\begin{tabular}{p{3.0cm}cccccccc}
\toprule
 & \multicolumn{3}{c}{Classification-only} & \multicolumn{3}{c}{Rotation-augmented} \\
\cmidrule(lr){2-4}
\cmidrule(lr){5-7}
Anomalous cars & MSP & ODIN & Accuracy & MSP & ODIN & Accuracy \\
\midrule
Model T & 26.77 $\pm$ 1.21 & 31.20 $\pm$ 1.22 & 72.92 $\pm$ 0.49 & 32.09 $\pm$ 0.86 & 36.10 $\pm$ 1.84 & 72.52 $\pm$ 0.62 \\
race car & 22.48 $\pm$ 2.53 & 27.12 $\pm$ 3.90 & 79.65 $\pm$ 1.85 & 20.32 $\pm$ 5.47 & 22.41 $\pm$ 6.41 & 74.67 $\pm$ 3.39 \\
sports car & 16.20 $\pm$ 1.77 & 13.86 $\pm$ 0.44 & 80.97 $\pm$ 1.97 & 16.80 $\pm$ 0.93 & 15.58 $\pm$ 0.96 & 81.00 $\pm$ 0.48 \\
minivan & 17.19 $\pm$ 2.57 & 17.78 $\pm$ 1.68 & 79.25 $\pm$ 1.89 & 17.32 $\pm$ 3.08 & 18.45 $\pm$ 2.67 & 80.01 $\pm$ 0.98 \\
ambulance & 11.13 $\pm$ 1.78 & 9.51 $\pm$ 0.97 & 75.71 $\pm$ 2.44 & 11.24 $\pm$ 0.84 & 10.61 $\pm$ 1.25 & 75.78 $\pm$ 0.31 \\
cab & 26.17 $\pm$ 2.42 & 27.93 $\pm$ 2.30 & 75.92 $\pm$ 3.77 & 28.57 $\pm$ 1.91 & 29.39 $\pm$ 2.52 & 76.74 $\pm$ 3.09 \\
beach wagon & 24.82 $\pm$ 0.85 & 26.30 $\pm$ 2.00 & 78.75 $\pm$ 1.09 & 24.50 $\pm$ 1.64 & 25.22 $\pm$ 1.89 & 79.81 $\pm$ 1.27 \\
jeep & 25.47 $\pm$ 0.38 & 26.99 $\pm$ 2.70 & 74.67 $\pm$ 1.37 & 27.92 $\pm$ 5.01 & 27.74 $\pm$ 3.28 & 72.84 $\pm$ 0.47 \\
convertible & 20.00 $\pm$ 2.63 & 18.35 $\pm$ 1.17 & 76.86 $\pm$ 0.81 & 15.32 $\pm$ 2.01 & 14.79 $\pm$ 2.24 & 76.26 $\pm$ 1.74 \\
limo & 25.16 $\pm$ 1.43 & 25.87 $\pm$ 0.83 & 77.04 $\pm$ 1.52 & 22.53 $\pm$ 2.08 & 23.49 $\pm$ 1.17 & 77.54 $\pm$ 1.19 \\
\midrule
Average & 21.54 $\pm$ 0.62 & 22.49 $\pm$ 0.54 & 77.17 $\pm$ 0.10 & 21.66 $\pm$ 0.19 & 22.38 $\pm$ 0.46 & 76.72 $\pm$ 0.19 \\
\bottomrule
\end{tabular}
\end{center}

\begin{center}
\centering
\scriptsize
\begin{tabular}{p{3.0cm}cccccccc}
\toprule
 & \multicolumn{3}{c}{Classification-only} & \multicolumn{3}{c}{Rotation-augmented} \\
\cmidrule(lr){2-4}
\cmidrule(lr){5-7}
Anomalous snakes & MSP & ODIN & Accuracy & MSP & ODIN & Accuracy \\
\midrule
ringneck snake & 20.18 $\pm$ 2.98 & 20.56 $\pm$ 2.78 & 71.08 $\pm$ 3.13 & 20.84 $\pm$ 0.77 & 23.22 $\pm$ 1.07 & 69.03 $\pm$ 0.46 \\
vine snake & 16.07 $\pm$ 4.51 & 17.19 $\pm$ 3.91 & 67.94 $\pm$ 3.96 & 15.94 $\pm$ 1.05 & 16.15 $\pm$ 0.96 & 72.65 $\pm$ 6.44 \\
hognose snake & 16.82 $\pm$ 0.38 & 16.65 $\pm$ 0.46 & 67.95 $\pm$ 2.81 & 19.70 $\pm$ 1.22 & 19.32 $\pm$ 0.77 & 69.85 $\pm$ 2.50 \\
thunder snake & 17.06 $\pm$ 2.94 & 19.18 $\pm$ 3.31 & 71.86 $\pm$ 7.58 & 21.26 $\pm$ 0.35 & 23.08 $\pm$ 0.70 & 69.34 $\pm$ 7.08 \\
garter snake & 21.45 $\pm$ 4.35 & 22.16 $\pm$ 3.81 & 67.26 $\pm$ 2.19 & 22.67 $\pm$ 1.13 & 23.12 $\pm$ 1.83 & 68.29 $\pm$ 5.90 \\
king snake & 17.37 $\pm$ 0.39 & 16.55 $\pm$ 1.19 & 66.45 $\pm$ 5.38 & 19.47 $\pm$ 3.72 & 17.96 $\pm$ 2.74 & 68.13 $\pm$ 2.74 \\
night snake & 21.70 $\pm$ 4.01 & 20.50 $\pm$ 3.36 & 76.56 $\pm$ 0.78 & 23.28 $\pm$ 0.79 & 24.12 $\pm$ 1.26 & 78.71 $\pm$ 3.91 \\
green snake & 12.42 $\pm$ 3.31 & 13.49 $\pm$ 3.57 & 71.07 $\pm$ 6.41 & 13.15 $\pm$ 1.11 & 13.94 $\pm$ 0.23 & 71.74 $\pm$ 2.75 \\
water snake & 24.50 $\pm$ 3.10 & 26.36 $\pm$ 3.24 & 67.46 $\pm$ 6.52 & 25.77 $\pm$ 3.59 & 29.62 $\pm$ 5.04 & 66.85 $\pm$ 0.70 \\
\midrule
Average & 18.62 $\pm$ 0.93 & 19.18 $\pm$ 0.79 & 69.74 $\pm$ 1.63 & 20.23 $\pm$ 0.18 & 21.17 $\pm$ 0.12 & 70.51 $\pm$ 0.48 \\
\bottomrule
\end{tabular}
\end{center}

\begin{center}
\centering
\scriptsize
\begin{tabular}{p{3.0cm}cccccccc}
\toprule
 & \multicolumn{3}{c}{Classification-only} & \multicolumn{3}{c}{Rotation-augmented} \\
\cmidrule(lr){2-4}
\cmidrule(lr){5-7}
Anomalous spiders & MSP & ODIN & Accuracy & MSP & ODIN & Accuracy \\
\midrule
tarantula & 19.45 $\pm$ 0.73 & 22.91 $\pm$ 2.37 & 60.67 $\pm$ 0.66 & 24.27 $\pm$ 3.25 & 26.07 $\pm$ 2.73 & 60.07 $\pm$ 0.86 \\
Argiope aurantia & 12.97 $\pm$ 0.39 & 12.49 $\pm$ 0.48 & 69.70 $\pm$ 2.17 & 12.82 $\pm$ 0.51 & 12.01 $\pm$ 0.21 & 69.17 $\pm$ 1.57 \\
barn spider & 23.03 $\pm$ 3.03 & 23.83 $\pm$ 2.95 & 75.69 $\pm$ 0.55 & 21.41 $\pm$ 1.41 & 23.54 $\pm$ 1.21 & 76.56 $\pm$ 1.85 \\
black widow & 29.24 $\pm$ 4.39 & 37.96 $\pm$ 5.68 & 61.79 $\pm$ 0.87 & 37.08 $\pm$ 7.50 & 42.64 $\pm$ 8.87 & 62.63 $\pm$ 0.09 \\
garden spider & 17.36 $\pm$ 1.33 & 15.51 $\pm$ 0.88 & 77.81 $\pm$ 2.29 & 16.57 $\pm$ 1.58 & 15.88 $\pm$ 1.42 & 76.38 $\pm$ 1.94 \\
wolf spider & 25.15 $\pm$ 2.98 & 32.23 $\pm$ 1.91 & 64.73 $\pm$ 0.45 & 25.48 $\pm$ 1.75 & 30.69 $\pm$ 1.28 & 67.19 $\pm$ 0.37 \\
\midrule
Average & 21.20 $\pm$ 0.56 & 24.15 $\pm$ 0.72 & 68.40 $\pm$ 0.21 & 22.90 $\pm$ 1.29 & 25.10 $\pm$ 1.78 & 68.68 $\pm$ 0.77 \\
\bottomrule
\end{tabular}
\end{center}

\begin{center}
\centering
\scriptsize
\begin{tabular}{p{3.0cm}cccccccc}
\toprule
 & \multicolumn{3}{c}{Classification-only} & \multicolumn{3}{c}{Rotation-augmented} \\
\cmidrule(lr){2-4}
\cmidrule(lr){5-7}
Anomalous fungi & MSP & ODIN & Accuracy & MSP & ODIN & Accuracy \\
\midrule
stinkhorn & 52.43 $\pm$ 1.15 & 56.37 $\pm$ 1.98 & 90.91 $\pm$ 0.54 & 54.37 $\pm$ 4.65 & 59.10 $\pm$ 5.71 & 92.27 $\pm$ 0.97 \\
bolete & 51.04 $\pm$ 0.42 & 52.82 $\pm$ 3.10 & 89.19 $\pm$ 0.94 & 49.43 $\pm$ 2.05 & 53.07 $\pm$ 3.48 & 89.22 $\pm$ 1.09 \\
hen-of-the-woods & 44.83 $\pm$ 1.52 & 48.04 $\pm$ 0.84 & 89.41 $\pm$ 1.64 & 48.87 $\pm$ 2.00 & 51.37 $\pm$ 2.44 & 90.13 $\pm$ 0.33 \\
earthstar & 34.90 $\pm$ 3.26 & 36.79 $\pm$ 2.16 & 86.70 $\pm$ 1.91 & 41.96 $\pm$ 7.66 & 43.24 $\pm$ 4.92 & 86.46 $\pm$ 0.62 \\
gyromitra & 46.75 $\pm$ 0.42 & 49.06 $\pm$ 2.64 & 86.79 $\pm$ 1.66 & 44.90 $\pm$ 1.94 & 49.20 $\pm$ 1.51 & 86.39 $\pm$ 0.18 \\
coral fungus & 25.42 $\pm$ 2.60 & 24.44 $\pm$ 3.04 & 86.36 $\pm$ 1.25 & 25.58 $\pm$ 1.15 & 25.22 $\pm$ 2.81 & 86.35 $\pm$ 0.80 \\
\midrule
Average & 42.56 $\pm$ 0.49 & 44.59 $\pm$ 1.46 & 88.23 $\pm$ 0.45 & 44.19 $\pm$ 1.86 & 46.86 $\pm$ 1.13 & 88.47 $\pm$ 0.43 \\
\bottomrule
\end{tabular}
\end{center}

%\clearpage

\section{Experiments with contrastive predictive coding (CPC) as the auxiliary task}
In this section, we provide further details of our experiments with CPC~\cite{cpc} as an auxiliary task. We only run these experiments on our proposed Imagenet subsets since as a patch-encoding predictive method, CPC has been developed primarily for signals with sufficient spatial or temporal dimensions for meaningful and sufficient subsampling. Existing work has explored the application to smaller images, but here we only focus on the most realistic and most difficult of the benchmarks we have proposed. \\ \\
CPC involves performing predictions for encodings of patches of an image from those above them. To avoid learning trivial codes, a contrastive loss is used which essentially trains the model to distinguish between correct codes and ``noisy'' ones. These negative samples are taken from patches within and across images in the batch. \\ \\
We use the same network architecture as we used for the Imagenet experiments with rotation-prediction as the auxiliary task, but modify the first convolution layer to have a stride of 2. This reduces the computational overhead sufficiently for concurrent training with CPC at reasonable batch-sizes (CPC training batch-sizes are 32), but at a minor expense of classification performance. We use the first three blocks of the network for the patch encoder as in~\cite{cpc}, and append the final layers for the classification task. Unlike with rotation, the auxiliary task works on patches while the primary classifier works on the entire image. This leads to differences in the operating receptive-fields, and differing proportions of boundary effects. To facilitate easier parameter sharing across the two tasks, we make the following changes. First, we replace all default zero-padding with symmetric-padding. This removes the effect of having a different ratio of border-zeros to pixels when the spatial dimensions of the input changes. Second, we replace all normalization layers with conditional normalization variants~\cite{de2017modulating}: this means separate sets of scale and shift parameters are used depending on the current predictive task. Since batch-normalization allows trivial solutions to CPC for patches sampled from different images as noted in~\cite{cpcplus}, we only use patches from within the same image, and find that we can continue using CPC to our advantage (although we found that such an implementation of CPC by itself leads to less linearly-separable representations compared to also taking negative samples from other images) . We keep the same optimizer settings from the rotation experiments, but it is possible that different choices might lead to further improvements. $\lambda$ is tuned to 10.0 for all experiments, following a coarse hyperparameter search for best validation-set classification accuracy over a range of $\{0.1, 1.0, 10.0, 20.0\}$. \\ \\
In the tables below, we show that similar patterns of improved anomaly detection and generalization are observed as with our experiments where rotation-prediction was the auxiliary task.
\vspace{1in}
\begin{center}
\centering
\scriptsize
\begin{tabular}{@{}p{3.0cm}cccccc@{}}
\toprule
 & \multicolumn{3}{c}{ Classification-only} & \multicolumn{3}{c}{CPC-augmented}  \\
\cmidrule(lr){2-4}
\cmidrule(lr){5-7}
Anomalous dog & MSP & ODIN & Accuracy & MSP & ODIN & Accuracy \\
\midrule
Ibizan hound & 20.07 $\pm$ 2.37 & 21.48 $\pm$ 3.04 & 84.08 $\pm$ 0.51 & 20.33 $\pm$ 1.12 & 21.23 $\pm$ 1.80 & 84.49 $\pm$ 0.62 \\
bluetick & 24.64 $\pm$ 3.66 & 30.79 $\pm$ 2.86 & 82.09 $\pm$ 1.06 & 27.23 $\pm$ 1.64 & 34.59 $\pm$ 1.10 & 83.86 $\pm$ 0.27 \\
beagle & 19.45 $\pm$ 0.62 & 18.49 $\pm$ 0.70 & 84.68 $\pm$ 0.45 & 20.17 $\pm$ 1.26 & 18.42 $\pm$ 1.03 & 85.56 $\pm$ 0.51 \\
Afghan hound & 17.46 $\pm$ 1.38 & 18.51 $\pm$ 1.17 & 80.95 $\pm$ 0.58 & 16.33 $\pm$ 1.48 & 20.60 $\pm$ 1.56 & 83.21 $\pm$ 0.84 \\
Weimaraner & 26.80 $\pm$ 5.68 & 32.76 $\pm$ 6.41 & 82.15 $\pm$ 0.78 & 26.09 $\pm$ 1.99 & 29.83 $\pm$ 2.19 & 82.72 $\pm$ 0.52 \\
Saluki & 25.22 $\pm$ 1.38 & 29.76 $\pm$ 0.66 & 82.96 $\pm$ 0.50 & 23.19 $\pm$ 0.74 & 27.07 $\pm$ 1.71 & 84.13 $\pm$ 0.13 \\
redbone & 16.47 $\pm$ 1.12 & 16.62 $\pm$ 1.68 & 83.78 $\pm$ 0.64 & 18.39 $\pm$ 2.35 & 17.91 $\pm$ 0.81 & 84.25 $\pm$ 0.16 \\
otterhound & 17.37 $\pm$ 1.80 & 17.41 $\pm$ 1.43 & 82.00 $\pm$ 0.81 & 16.37 $\pm$ 1.22 & 16.44 $\pm$ 1.25 & 83.77 $\pm$ 0.21 \\
Norweigian elkhound & 23.71 $\pm$ 3.89 & 27.66 $\pm$ 3.81 & 81.26 $\pm$ 0.17 & 28.82 $\pm$ 2.58 & 38.19 $\pm$ 4.92 & 82.53 $\pm$ 0.59 \\
basset hound & 18.63 $\pm$ 1.45 & 17.53 $\pm$ 1.87 & 84.59 $\pm$ 0.62 & 18.04 $\pm$ 1.24 & 17.11 $\pm$ 0.64 & 85.36 $\pm$ 0.30 \\
Scottish deerhound & 21.70 $\pm$ 0.23 & 19.70 $\pm$ 0.67 & 82.79 $\pm$ 0.79 & 23.23 $\pm$ 2.82 & 23.42 $\pm$ 3.56 & 83.46 $\pm$ 0.55 \\
bloodhound & 18.53 $\pm$ 1.63 & 22.59 $\pm$ 2.13 & 86.49 $\pm$ 0.72 & 18.96 $\pm$ 1.09 & 24.09 $\pm$ 1.99 & 86.55 $\pm$ 1.02 \\
\midrule
Average & 20.84 $\pm$ 0.50 & 22.77 $\pm$ 0.74 & 83.12 $\pm$ 0.26 & 21.43 $\pm$ 0.63 & 24.08 $\pm$ 0.63 & 84.16 $\pm$ 0.07 \\
\bottomrule
\end{tabular}
\end{center}

\begin{center}
\centering
\scriptsize
\begin{tabular}{@{}p{3.0cm}cccccc@{}}
\toprule
 & \multicolumn{3}{c}{ Classification-only} & \multicolumn{3}{c}{CPC-augmented}  \\
\cmidrule(lr){2-4}
\cmidrule(lr){5-7}
Anomalous car & MSP & ODIN & Accuracy & MSP & ODIN & Accuracy \\
\midrule
Model T & 24.48 $\pm$ 2.28 & 27.42 $\pm$ 2.11 & 71.00 $\pm$ 0.98 & 27.08 $\pm$ 2.01 & 30.44 $\pm$ 2.39 & 76.90 $\pm$ 0.60 \\
race car & 21.22 $\pm$ 0.78 & 26.82 $\pm$ 1.07 & 76.79 $\pm$ 0.80 & 20.19 $\pm$ 2.57 & 24.54 $\pm$ 3.58 & 80.22 $\pm$ 0.16 \\
sports car & 15.16 $\pm$ 0.82 & 14.34 $\pm$ 0.63 & 79.40 $\pm$ 0.47 & 19.33 $\pm$ 1.45 & 15.33 $\pm$ 0.96 & 81.70 $\pm$ 0.53 \\
minivan & 16.92 $\pm$ 2.16 & 18.60 $\pm$ 2.95 & 77.48 $\pm$ 0.58 & 17.87 $\pm$ 0.46 & 19.54 $\pm$ 2.34 & 80.38 $\pm$ 0.62 \\
ambulance & 11.12 $\pm$ 0.58 & 10.36 $\pm$ 0.28 & 73.16 $\pm$ 0.40 & 11.41 $\pm$ 2.43 & 11.32 $\pm$ 2.66 & 76.70 $\pm$ 1.36 \\
cab & 23.52 $\pm$ 1.33 & 27.08 $\pm$ 1.13 & 76.26 $\pm$ 0.65 & 26.45 $\pm$ 1.01 & 28.11 $\pm$ 1.64 & 78.51 $\pm$ 0.38 \\
beach wagon & 23.52 $\pm$ 2.35 & 23.60 $\pm$ 0.96 & 76.82 $\pm$ 0.34 & 24.57 $\pm$ 0.68 & 27.34 $\pm$ 1.32 & 80.96 $\pm$ 0.50 \\
jeep & 25.72 $\pm$ 0.32 & 27.31 $\pm$ 0.89 & 73.10 $\pm$ 0.85 & 27.37 $\pm$ 3.33 & 29.21 $\pm$ 2.51 & 76.77 $\pm$ 0.63 \\
convertible & 15.05 $\pm$ 0.51 & 14.11 $\pm$ 0.60 & 74.80 $\pm$ 0.19 & 20.84 $\pm$ 2.61 & 21.43 $\pm$ 2.43 & 78.84 $\pm$ 0.85 \\
limo & 21.91 $\pm$ 2.30 & 24.52 $\pm$ 2.90 & 75.38 $\pm$ 0.63 & 26.97 $\pm$ 2.27 & 28.84 $\pm$ 2.50 & 77.84 $\pm$ 0.97 \\
\midrule
Average & 19.86 $\pm$ 0.21 & 21.42 $\pm$ 0.48 & 75.42 $\pm$ 0.11 & 22.21 $\pm$ 0.44 & 23.61 $\pm$ 0.57 & 78.88 $\pm$ 0.15 \\
\bottomrule
\end{tabular}
\end{center}

\begin{center}
\centering
\scriptsize
\begin{tabular}{@{}p{3.0cm}cccccc@{}}
\toprule
 & \multicolumn{3}{c}{ Classification-only} & \multicolumn{3}{c}{CPC-augmented}  \\
\cmidrule(lr){2-4}
\cmidrule(lr){5-7}
Anomalous snake & MSP & ODIN & Accuracy & MSP & ODIN & Accuracy \\
\midrule
ringneck snake & 19.14 $\pm$ 0.52 & 19.89 $\pm$ 1.12 & 63.49 $\pm$ 3.27 & 20.06 $\pm$ 1.20 & 22.98 $\pm$ 0.70 & 66.64 $\pm$ 0.51 \\
vine snake & 13.91 $\pm$ 1.83 & 15.51 $\pm$ 2.13 & 64.52 $\pm$ 2.01 & 14.81 $\pm$ 1.27 & 15.63 $\pm$ 0.45 & 67.14 $\pm$ 2.57 \\
hognose snake & 15.07 $\pm$ 0.84 & 13.72 $\pm$ 0.45 & 65.56 $\pm$ 3.19 & 13.73 $\pm$ 0.93 & 14.10 $\pm$ 1.15 & 67.52 $\pm$ 2.72 \\
thunder snake & 19.16 $\pm$ 0.03 & 19.35 $\pm$ 0.92 & 70.29 $\pm$ 2.43 & 20.25 $\pm$ 0.71 & 20.07 $\pm$ 2.81 & 72.44 $\pm$ 4.05 \\
garter snake & 20.86 $\pm$ 2.26 & 22.62 $\pm$ 2.94 & 63.59 $\pm$ 2.32 & 20.25 $\pm$ 2.03 & 23.80 $\pm$ 2.18 & 65.23 $\pm$ 2.32 \\
king snake & 17.36 $\pm$ 1.34 & 15.17 $\pm$ 2.04 & 62.43 $\pm$ 2.04 & 19.35 $\pm$ 2.62 & 18.16 $\pm$ 2.95 & 64.61 $\pm$ 2.39 \\
night snake & 20.67 $\pm$ 0.10 & 19.81 $\pm$ 1.18 & 73.04 $\pm$ 2.02 & 21.94 $\pm$ 2.49 & 22.78 $\pm$ 0.85 & 73.62 $\pm$ 2.60 \\
green snake & 12.31 $\pm$ 0.63 & 13.63 $\pm$ 0.90 & 67.11 $\pm$ 4.65 & 12.43 $\pm$ 1.73 & 12.92 $\pm$ 0.97 & 65.58 $\pm$ 2.37 \\
water snake & 25.30 $\pm$ 2.04 & 28.33 $\pm$ 3.52 & 65.35 $\pm$ 4.95 & 26.20 $\pm$ 2.50 & 33.06 $\pm$ 2.34 & 69.38 $\pm$ 3.72 \\
\midrule
Average & 18.20 $\pm$ 0.76 & 18.67 $\pm$ 1.07 & 66.15 $\pm$ 1.89 & 18.78 $\pm$ 0.40 & 20.39 $\pm$ 0.60 & 68.02 $\pm$ 0.85 \\
\bottomrule
\end{tabular}
\end{center}

\begin{center}
\centering
\scriptsize
\begin{tabular}{@{}p{3.0cm}cccccc@{}}
\toprule
 & \multicolumn{3}{c}{ Classification-only} & \multicolumn{3}{c}{CPC-augmented}  \\
\cmidrule(lr){2-4}
\cmidrule(lr){5-7}
Anomalous spider & MSP & ODIN & Accuracy & MSP & ODIN & Accuracy \\
\midrule
tarantula & 22.42 $\pm$ 2.32 & 22.47 $\pm$ 2.51 & 58.71 $\pm$ 1.58 & 21.94 $\pm$ 0.22 & 23.34 $\pm$ 1.48 & 61.46 $\pm$ 0.66 \\
Argiope aurantia & 13.84 $\pm$ 0.37 & 12.98 $\pm$ 0.38 & 66.66 $\pm$ 1.04 & 14.44 $\pm$ 0.95 & 12.93 $\pm$ 0.34 & 68.59 $\pm$ 1.97 \\
barn spider & 25.39 $\pm$ 1.40 & 25.81 $\pm$ 2.52 & 74.17 $\pm$ 0.96 & 20.60 $\pm$ 2.90 & 22.92 $\pm$ 3.42 & 75.76 $\pm$ 1.91 \\
black widow & 24.20 $\pm$ 1.60 & 29.34 $\pm$ 3.18 & 60.60 $\pm$ 1.28 & 28.93 $\pm$ 0.91 & 34.29 $\pm$ 1.42 & 63.52 $\pm$ 1.21 \\
garden spider & 17.21 $\pm$ 0.28 & 16.02 $\pm$ 0.20 & 75.05 $\pm$ 1.57 & 17.90 $\pm$ 0.53 & 16.65 $\pm$ 0.19 & 75.87 $\pm$ 1.79 \\
wolf spider & 29.11 $\pm$ 2.91 & 37.87 $\pm$ 2.64 & 64.72 $\pm$ 1.38 & 29.88 $\pm$ 0.83 & 30.10 $\pm$ 0.33 & 66.79 $\pm$ 1.74 \\
\midrule
Average & 22.03 $\pm$ 0.68 & 24.08 $\pm$ 0.70 & 66.65 $\pm$ 0.42 & 22.28 $\pm$ 0.60 & 23.37 $\pm$ 0.68 & 68.67 $\pm$ 0.36 \\
\bottomrule
\end{tabular}
\end{center}

\begin{center}
\centering
\scriptsize
\begin{tabular}{@{}p{3.0cm}cccccc@{}}
\toprule
 & \multicolumn{3}{c}{ Classification-only} & \multicolumn{3}{c}{CPC-augmented}  \\
\cmidrule(lr){2-4}
\cmidrule(lr){5-7}
Anomalous fungus & MSP & ODIN & Accuracy & MSP & ODIN & Accuracy \\
\midrule
stinkhorn & 46.05 $\pm$ 1.98 & 51.30 $\pm$ 1.59 & 89.81 $\pm$ 0.78 & 50.69 $\pm$ 3.13 & 58.81 $\pm$ 5.49 & 91.27 $\pm$ 0.08 \\
bolete & 46.73 $\pm$ 2.58 & 50.28 $\pm$ 6.04 & 88.41 $\pm$ 0.42 & 49.19 $\pm$ 3.72 & 51.67 $\pm$ 2.73 & 90.87 $\pm$ 0.43 \\
hen-of-the-woods & 43.58 $\pm$ 2.47 & 47.98 $\pm$ 1.84 & 88.10 $\pm$ 0.45 & 38.97 $\pm$ 2.87 & 42.59 $\pm$ 2.33 & 90.63 $\pm$ 0.58 \\
earthstar & 35.63 $\pm$ 0.71 & 36.72 $\pm$ 1.59 & 84.75 $\pm$ 0.88 & 39.83 $\pm$ 3.16 & 40.01 $\pm$ 3.92 & 85.42 $\pm$ 0.75 \\
gyromitra & 39.90 $\pm$ 1.59 & 42.04 $\pm$ 2.19 & 86.35 $\pm$ 0.45 & 45.44 $\pm$ 2.50 & 49.57 $\pm$ 0.86 & 87.99 $\pm$ 0.85 \\
coral fungus & 23.25 $\pm$ 2.15 & 21.93 $\pm$ 2.50 & 84.89 $\pm$ 1.16 & 28.35 $\pm$ 0.43 & 27.67 $\pm$ 3.76 & 87.29 $\pm$ 0.62 \\
\midrule
Average & 39.19 $\pm$ 1.26 & 41.71 $\pm$ 1.94 & 87.05 $\pm$ 0.06 & 42.08 $\pm$ 0.57 & 45.05 $\pm$ 1.11 & 88.91 $\pm$ 0.46 \\
\bottomrule
\end{tabular}
\end{center}

%\clearpage

\section{Trivial baseline for OOD detection on existing benchmarks}
To demonstrate that the current benchmarks are trivial with very low-level information, we tested OOD detection with \cifar as the in-distribution by simply looking at likelihoods under a mixture of 3 Gaussians, trained channel-wise at a pixel-level. We find that this simple baseline compares very well with approaches in recent papers at all but one of the benchmark OOD tasks in~\cite{odin} for \cifar, as we show below:

\begin{center}
\centering
\begin{tabular}{p{4cm}c}
   \toprule
   OOD dataset & Average precision \\
   \midrule
   TinyImagenet (crop)  & 96.84 \\
   TinyImagenet (resize) & 99.03 \\
   LSUN & 58.06 \\
   LSUN (resize) & 99.77 \\
   iSUN & 99.21 \\
   \bottomrule
\end{tabular}
\end{center}
We see that this method does not do well on LSUN. When we inspect LSUN, we find that the images are cropped patches from scene-images, and a majority of them are of uniform colour and texture, with little variation and structure in them. While this dataset is most obviously different from the in-distribution examples from \cifar, we believe that the particular appearance of the images results in the phenomenon reported in~\cite{glowfail}, where one distribution that ``sits inside'' the other because of a similar mean but lower variance ends up being more likely under the wider distribution. In fact, thresholding on simply the ``energy'' of the edge-detection map gives us an average precision of around 87.5\% for LSUN, thus indicating that the extremely trivial feature of a lower edge-count is already a strong indicator for telling apart such an obvious difference. \\ \\
We found that this simple baseline of pixel-level channel-mixture of Gaussians underperforms severely on the hold-out-class experiments on CIFAR-10, achieving an average precision of a mere 11.17\% across the 10 experiments.

\end{document}